\documentclass[]{servicenow}

\usepackage[utf8]{inputenc}
\usepackage[T1]{fontenc}
\usepackage[inline]{enumitem}
\usepackage{graphicx}
\usepackage{url}
\usepackage{subcaption}
\usepackage{nicefrac}
\usepackage[table]{xcolor}
\usepackage{array}
\usepackage{booktabs}
\usepackage{amsfonts}
\usepackage{amsmath}
\usepackage{amssymb}
\usepackage{microtype}
\usepackage{makecell}
\usepackage{colortbl}
\usepackage{empheq}
\usepackage{tcolorbox}
\usepackage{bbding}
\usepackage{tabularx}
\usepackage{multirow}
\usepackage{wrapfig2}
\usepackage{bbm}
\usepackage{pifont}
\usepackage{color}

\usepackage{hyperref}   
\usepackage{cleveref}   

\usepackage{silence}
\definecolor{dkgreen}{rgb}{0,0.6,0}
\definecolor{gray}{rgb}{0.5,0.5,0.5}
\definecolor{mauve}{rgb}{0.58,0,0.82}

\definecolor{dgreen}{rgb}{0.412,0.741,0.271}
\definecolor{dblue}{rgb}{0.220,0.325,0.639}
\definecolor{dred}{rgb}{0.933,0.122,0.137}

\definecolor{memhighlight}{RGB}{227,242,253}
\definecolor{memframe}{RGB}{21,101,192}
\definecolor{abstainhighlight}{RGB}{255,243,224}
\definecolor{abstainframe}{RGB}{230,81,0}

\definecolor{bestcell}{RGB}{232,245,233}    
\definecolor{rowshade}{RGB}{248,248,248}    
\definecolor{csframe}{RGB}{46,125,50}       
\definecolor{csbg}{RGB}{250,253,250}        
\definecolor{memcolor}{RGB}{25,118,210}     
\definecolor{abstaincolor}{RGB}{239,108,0}  
\definecolor{darkgreen}{RGB}{27,94,32}
\definecolor{wacell}{RGB}{232,240,254}      
\definecolor{labcell}{RGB}{255,243,224}     

\definecolor{g1}{HTML}{b3e2cd}
\definecolor{r1}{HTML}{fdcdac}
\definecolor{w1}{HTML}{cbd5e8}
\definecolor{b1}{HTML}{fff7bc}

\definecolor{lr}{HTML}{bebada}
\definecolor{fr}{HTML}{fccde5}

\definecolor{Lavender}{HTML}{BF94E4}

\newcommand{\ie}{\textit{i}.\textit{e}.\,}
\newcommand{\eg}{\textit{e}.\textit{g}.\,}

\definecolor{l1}{RGB}{189,215,238}
\definecolor{l2}{RGB}{222,235,247}
\definecolor{l3}{RGB}{255,230,153}
\definecolor{l4}{RGB}{248,203,173}
\definecolor{l5}{RGB}{244,177,131}

\newtcbox{\columntcbox}{highlight math style={
        colback=gray!30,
        arc=2pt,
        outer arc=2pt,
        boxrule=0pt,
        top=2pt,
        bottom=2pt,
        left=2pt,
        right=2pt,
    }
}

\colorlet{LightLavender}{Lavender!35!}
\newtcbox{\inlinetcbox}[1][]{on line, 
        boxsep=2pt, left=0pt,right=0pt,top=0pt,bottom=0pt,
        colframe=white,colback=LightLavender,  
        highlight math style={enhanced}, #1
}

\newcolumntype{C}[1]{>{\centering}m{#1}}

\makeatletter
\newcommand{\xMapsto}[2][]{\ext@arrow 0599{\Mapstofill@}{#1}{#2}}
\def\Mapstofill@{\arrowfill@{\Mapstochar\Relbar}\Relbar\Rightarrow}
\makeatother

\newcommand{\ourmethod}{\texorpdfstring{\texttt{Mem-$\pi$}}{Mem-pi}}
\newcommand{\mempolicy}{\ensuremath{\pi_\text{mem}}}
\newcommand{\mempolicyone}{\ensuremath{\pi_\text{mem}^{1}}}
\newcommand{\mempolicytwo}{\ensuremath{\pi_\text{mem}^{2}}}
\newcommand{\agentpolicy}{\ensuremath{\pi_\text{agent}}}

\title{\ourmethod{}: Adaptive Memory through Learning \\
       When and What to Generate}

\author[1,2,3]{Xiaoqiang Wang}
\author[1]{Chao Wang}
\author[1,2,3]{Hadi Nekoei}
\author[1,2,4,6]{Christopher Pal}
\author[1]{Alexandre Lacoste}
\author[1,5]{Spandana Gella}
\author[2,3,6]{Bang Liu}
\author[1,2,5]{Perouz Taslakian}

\affiliation[1]{ServiceNow AI Research}
\affiliation[2]{Mila -- Quebec AI Institute}
\affiliation[3]{Universit\'e de Montr\'eal}
\affiliation[4]{Polytechnique Montr\'eal}
\affiliation[5]{McGill University}
\affiliation[6]{CIFAR AI Chair}

\correspondence{\email{bang.liu@umontreal.ca}, \email{perouz.taslakian@servicenow.com}}

\abstract{%
We present \ourmethod{}, a framework for adaptive memory in large language
model (LLM) agents, where useful guidance is \emph{generated on demand}
rather than retrieved from external memory stores. Existing memory-augmented
agents typically rely on similarity-based retrieval from episodic memory
banks or skill libraries, returning static entries that often misalign with
the current query context. In contrast, we model memory as a generative
policy realized by a dedicated language or vision-language model with its
own parameters, separate from the downstream agent, and fine-tuned
specifically to produce context-specific guidance that cues the agent on
how to perform complex tasks. The memory policy jointly decides \emph{when}
to produce guidance and \emph{what} guidance to produce, trained with a
decision-content decoupled reinforcement learning (RL) objective so that it
abstains when generation would not help and otherwise produces concise,
task-relevant guidance. Across diverse agentic benchmarks spanning web
navigation, terminal tool use, and embodied environments, \ourmethod{}
consistently outperforms retrieval-based and RL-optimized memory baselines,
achieving over 20\% relative improvement on average.
}

\begin{document}
\maketitle

\section{Introduction}
\label{sec:introduction}

\begin{figure}[!t]
\centering
    \includegraphics[width=1.0\textwidth]{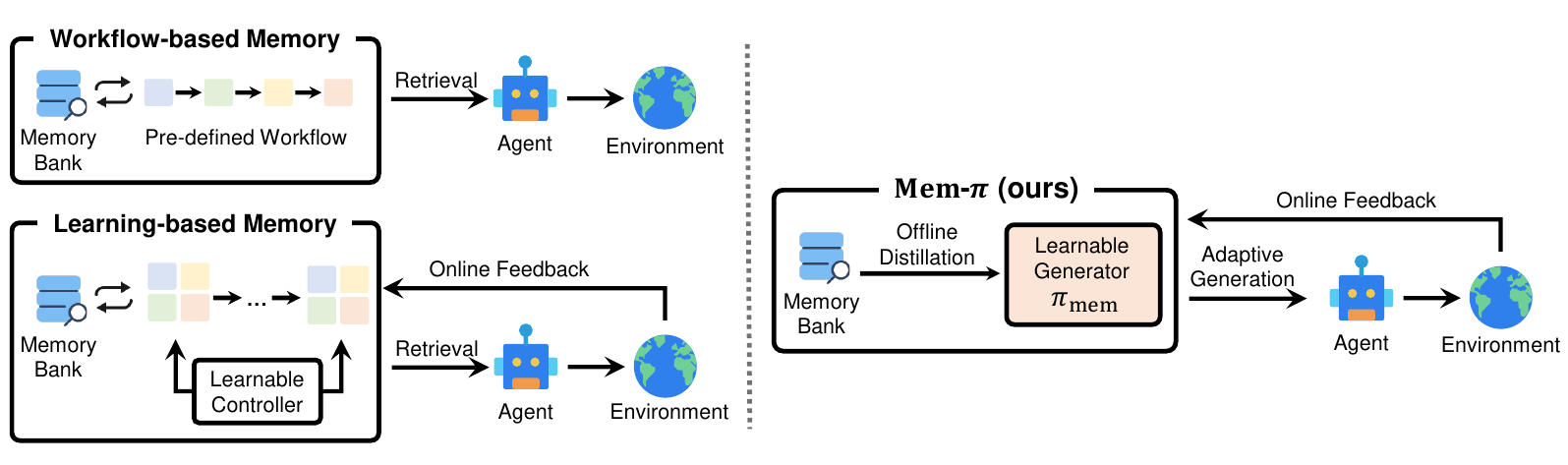}
    \caption{
        Comparison of (a) workflow-based memory systems, where memory operations are governed by predefined retrieval and update pipelines, (b) learning-based memory systems, where memory operations are jointly optimized with downstream agent outcomes, and (c) our \ourmethod{}, which models memory as a generative policy \mempolicy{} separate from the downstream agent and internalizes reusable experience through offline experience distillation and online adaptation distillation.
    }
    \label{fig:motivation-comparison}
    \vspace{-5mm}
\end{figure}

Large language models (LLMs)~\citep{ouyang2022training,team2023gemini,hurst2024gpt,deepseekai2026deepseekv4} have demonstrated remarkable capabilities on reasoning-intensive benchmarks~\citep{liang2022holistic,srivastava2023beyond,phan2025humanity,deng2025swe} and shown potential as autonomous agents~\citep{liu2025advances} operating in real-world environments, enabling applications such as computer-use agents~\citep{wang2025oscar,qin2025ui,zhang2025appagent}, deep research assistants~\citep{openai2025deepresearch,li2025webthinker,han2025deep}, and automated scientific discovery systems~\citep{lu2024ai,schmidgall-etal-2025-agent,liu2026evox}.
Despite these advances, current LLMs remain limited by their stateless nature and cannot accumulate reusable experience across interactions~\citep{sumers2023cognitive,tao2024survey}.
To address this limitation, recent agent systems augment LLMs with external memory modules~\citep{zhang2025survey,huang2026rethinking,zhou2026externalization}, such as episodic memory banks~\citep{zhong2024memorybank,cai2025flex} and reusable skill libraries~\citep{wang2023voyager,xia2026skillrl,shi2026evolving} distilled from prior interactions (Figure~\ref{fig:motivation-comparison}).

Existing memory-augmented agents collect memory fragments into a bank and retrieve relevant entries at inference time.
Early approaches use \emph{workflow-based memory}~\citep{packer2023memgpt,fu2024autoguide,ouyang2025reasoningbank}, where predefined rules govern memory construction, retrieval, and update~\citep{zhao2024expel,wang2025agent}.
Recent work explores \emph{learning-based memory}~\citep{yan2025memory,zhou2025memento,zhang2025memevolve,zhang2026memrl}, optimizing memory operations end-to-end via downstream task outcomes.
However, both lines remain constrained by a retrieval-based paradigm that reuses explicitly stored experiences. Retrieved memories are inherently static and often contain irrelevant~\citep{wang-etal-2024-crafting,xu2026chain}, partially aligned, or overly specific information~\citep{yang2026beyond} that cannot adapt to the agent's current context.

Cognitive science suggests a different view: human remembering is not a literal replay mechanism~\citep{nosofsky1994rule,ashby2011human} but a \emph{constructive} process, where recollection is dynamically reconstructed from prior knowledge and the current context~\citep{bartlett1932remembering,schacter1998cognitive,schacter2007cognitive}.
%
%
Concurrent work such as ParaMem~\citep{yao2026parammem} and SEAM~\citep{li2026beyond} replaces retrieved memory with generated memory~\citep{wang-etal-2025-r3mem,wu2025towards,zhang2025memgen}, but either applies it without conditioning on the current context or invokes generation as an always-on auxiliary step.
This raises a reliability concern that is especially acute in agent settings, where memory is not the final task output but an intervention on a downstream agent. Under ambiguous, weakly grounded, or out-of-distribution contexts, generated guidance can be uninformative or even harmful, propagating hallucinated cues into agent actions.

Building on this, we present \ourmethod{}, a framework for adaptive memory generation in LLM agents.
Rather than retrieving fixed entries or unconditionally generating auxiliary guidance, \ourmethod{} models memory as a parametric policy \mempolicy{} that learns both \emph{when} to generate and \emph{what} to generate.
Conditioned on the agent context (\ie, task instructions and environment observations), \ourmethod{} produces concise, task-adaptive guidance from reusable experience encoded in its parameters.

Encoding experience in an \ourmethod{} model's parameters brings several advantages. 
First, its memory footprint is bounded by model size rather than the number of accumulated experiences, reducing the growing memory-management overhead associated with merging~\citep{yin-etal-2024-explicit,hu2024hiagent} and forgetting~\citep{zhong2024memorybank}.
Second, since \mempolicy{} synthesizes guidance on demand rather than copying stored entries, it can fuse signals from many past experiences into a single context-specific hint, unlike top-$k$ retrieval, which may split them across fragments or omit them beyond the cutoff~\citep{jiang-etal-2023-active,asai2023self,jeong-etal-2024-adaptive}.
Finally, this framework separates specialization from execution: a smaller private local model can be fine-tuned as \mempolicy{} and plugged into a larger or frontier agent model to leverage broader reasoning capabilities.


We train \ourmethod{} in two stages. \emph{Experience distillation} first compresses an offline experience bank into the memory policy via supervised learning, internalizing reusable behaviors.
\emph{Adaptation distillation} then refines the policy through reinforcement learning, using downstream task outcomes as the reward signal to align memory generation with task success.
To ensure reliability, we incorporate \emph{abstention} into \mempolicy{}, allowing it to skip memory generation when generation is unnecessary or uncertain.
Specifically, we introduce a decision-content decoupled objective built on Group Relative Policy Optimization (GRPO)~\citep{shao2024deepseekmath} that separates \emph{when} to generate from \emph{what} to generate.
The objective uses structured counterfactual rollouts to compare the two branches, decomposing learning into decision-level and content-level advantages and enabling adaptive memory behavior: the policy generates guidance only when it improves downstream task outcomes, and abstains otherwise.

We evaluate \ourmethod{} across diverse agent benchmarks spanning web navigation (\textsc{WebArena}~\citep{zhou2023webarena}, \mbox{\textsc{WorkArena}~\citep{drouin2024workarena})}, terminal tool use (\textsc{LifelongAgentBench}~\citep{zheng2025lifelongagentbench}), and text-based embodied environments (\textsc{ALFWorld}~\citep{shridhar2020alfworld}). 
Adaptive memory generation consistently outperforms retrieval-based memory baselines across all four benchmarks, yielding a 20\% relative gain over the base agent on average, with the relative gain on \textsc{WebArena} approaching 50\%.

\begin{figure}[!t]
\centering
    \includegraphics[width=0.95\textwidth]{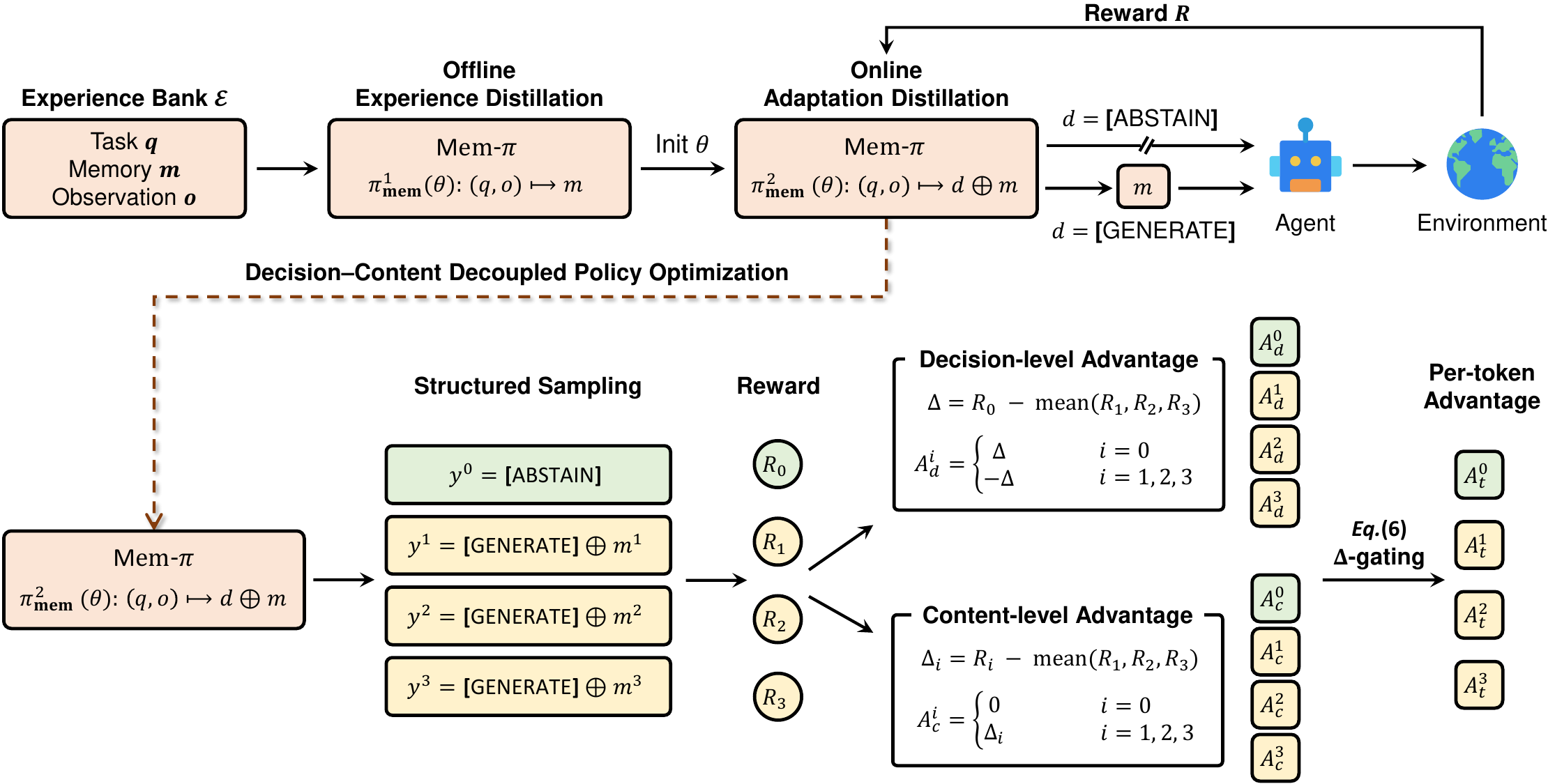}
    \caption{
        Overview of \ourmethod{}.
        We train the generative memory policy \mempolicy{} in two stages.
        \emph{Experience Distillation} distills reusable experience from an offline experience bank via supervised learning.
        \emph{Adaptation Distillation} then optimizes \mempolicy{} against downstream agent outcomes using decision-content decoupled policy optimization, which pairs abstain and generate rollouts and decomposes the GRPO advantage into decision-level and content-level signals.
    }
    \label{fig:training-pipeline}
    \vspace{-5mm}
\end{figure}

\section{Design of \ \ourmethod{}}
\label{sec:methodology}

We model adaptive memory as a generative policy \mempolicy{} parameterized by $\theta$ and instantiated as a dedicated language or vision-language model \ourmethod{}, separate from the downstream agent. \ourmethod{} produces guidance that is injected into the agent's context at inference time. Let $\mathcal{E}$ denote an offline experience bank of context-guidance pairs $(x,m)$, where each task context $x=(q,o)$ consists of a task specification $q$ and an environment observation $o\in\mathcal{O}$, and each memory guidance $m\in\mathcal{M}$ is a textual hint inserted into the downstream agent's context to inform its decisions. The observation $o$ may include structured textual representations and, when available, visual inputs such as screenshots in web navigation tasks. Figure~\ref{fig:bank} illustrates the structure of each field.

First, \emph{experience distillation} learns a mapping $\mempolicyone{}:(q,o)\mapsto m$ via supervised learning on $\mathcal{E}$, converting explicit offline experiences into parametric knowledge so that the policy can produce context-specific guidance for new tasks at inference time. This design is inspired by context-supervised pretraining~\citep{gao-callan-2022-unsupervised,w-etal-2023-query}, where models learn to reconstruct knowledge from context and internalize it into their parameters. Let $m_t$ denote the $t$-th token of the target memory $m$, and let $m_{<t}$ denote its preceding tokens. We optimize \mempolicyone{} with the autoregressive supervised objective:
\begin{equation}
\mathcal{J}_{\mathrm{mem}}^{(1)}(\theta)
=
\mathbb{E}_{(x,m)\sim\mathcal{E}}
\left[
\sum_{t=1}^{|m|}
\log \mempolicyone{}\!\left(m_t \mid x, m_{<t} ; \theta\right)
\right].
\label{eq:stage1-objective}
\end{equation}
Second, \emph{adaptation distillation} (Section~\ref{sec:adaptation-distillation}) initializes \mempolicytwo{} from \mempolicyone{} and further optimizes the shared parameters $\theta$ through reinforcement learning from downstream agent outcomes, aligning memory generation with task utility rather than imitation quality alone.
To support reliable memory use, we introduce an explicit abstention decision, enabling the policy to skip generation when guidance is unnecessary or potentially unhelpful.
Specifically, we extend the output space with a decision token and define the mapping $\mempolicytwo{}:(q,o)\mapsto y$, where $y=d\oplus m$, $d\in\{\texttt{[GENERATE]},\texttt{[ABSTAIN]}\}$, $m\in\mathcal{M}\cup\{\varnothing\}$, and $\oplus$ denotes string concatenation.
When $d=\texttt{[GENERATE]}$, the policy emits guidance $m\in\mathcal{M}$, which is prepended to the downstream agent input to form the augmented context $x\oplus m$. When $d=\texttt{[ABSTAIN]}$, we set $m=\varnothing$, and the agent operates on the original context $x$.

A key challenge in Stage 2 is the imbalance between the routing decision and the memory content. The decision is encoded by a short token prefix, whereas the generated guidance spans a much longer sequence. As a result, under a flat policy-gradient objective, content-level gradients can dominate decision-level learning. We introduce a \emph{decision-content decoupled} objective (Section~\ref{sec:decoupled-optimization}) that separates routing and content learning signals through decomposing flat advantage.

\subsection{Adaptation Distillation}
\label{sec:adaptation-distillation}

While experience distillation provides a strong initialization, the supervised policy cannot determine \emph{when} memory generation is useful or potentially harmful.
Moreover, its guidance remains bounded by the offline experience bank and is not directly optimized for the needs of the downstream agent.
The adaptation distillation addresses this by refining $\pi_\text{mem}$ with RL using agent outcomes as the reward signal.
We extend the model vocabulary with two special tokens, \ie, decision tokens including \texttt{[GENERATE]} and \texttt{[ABSTAIN]}, and initialize their embeddings symmetrically so that both decisions have comparable initial probabilities and can be sufficiently explored at the beginning of training.

We adopt GRPO~\citep{shao2024deepseekmath} as the base RL algorithm, which removes the need for value models by estimating advantages from grouped samples.
For each $x$, GRPO samples a group of $G$ outputs $\{y^1,\ldots,y^G\}$ from $\pi_\text{mem}$ and computes group-relative advantages \mbox{$\hat{A}^j = (r^j-\operatorname{mean}(\mathbf{r}))/(\operatorname{std}(\mathbf{r})+\epsilon_{\mathrm{std}})$}, where $\mathbf{r}=(r^1,\ldots,r^G)$. The policy is updated by maximizing: 
\begin{equation}
\mathcal{J}_\text{GRPO}(\theta) = \mathbb{E}_{x} \!\left[ \frac{1}{G} \sum_{j=1}^{G} \frac{1}{|y^j|} \sum_{t=1}^{|y^j|} \!\left( \min\!\Big( \rho_t^j \hat{A}^j,\; \operatorname{clip}\!\big(\rho_t^j, 1\!-\!\epsilon_{\mathrm{clip}}, 1\!+\!\epsilon_{\mathrm{clip}}\big) \hat{A}^j \Big) - \beta \, D_{\mathrm{KL}}^{(t)} \right) \right]\!,
\label{eq:grpo-objective}
\end{equation}
where $\rho_t^j =
\pi_\text{mem}^{2}(y_t^j \mid x, y_{<t}^j; \theta)
/ \pi_\text{old}(y_t^j \mid x, y_{<t}^j; \theta_\text{old})$
is the importance ratio between the current policy and the old policy used for rollout sampling, and $D_{\mathrm{KL}}^{(t)}$ denotes the token-level KL divergence from a reference policy $\pi_\text{ref}$.
Here, $\pi_\text{old}$ is a frozen snapshot of \mempolicytwo{}, and $\pi_{\mathrm{ref}}$ is set to the Stage-1 policy \mempolicyone{} before adaptation distillation.


\noindent \textbf{Reward design.}
The reward $r=R(x,y)$ consists of a downstream task reward and, for generated memories, a length regularizer $R_m$. Given $y=d\oplus m$, we define:
\begin{equation}
R(x,y) =
\begin{cases}
\operatorname{TaskReward}\!\big(\agentpolicy{}(\cdot \mid x \oplus m)\big) + R_m(m),
& \text{if } d=\texttt{[GENERATE]} \\[2pt]
\operatorname{TaskReward}\!\big(\agentpolicy{}(\cdot \mid x)\big),
& \text{if } d=\texttt{[ABSTAIN]},
\end{cases}
\label{eq:reward}
\end{equation}
where \agentpolicy{} denotes the downstream agent's action distribution, which is not trained in this stage, and $\operatorname{TaskReward}(\cdot)\in\{0,1\}$ is a binary signal indicating task success or failure from the agent's interaction trajectory under the memory-augmented or original context.
Following length-aware reward shaping in reasoning and agentic LLMs~\citep{aggarwal2025l1,yu2025dapo,liu2025dler}, we use $R_m(m) = -\lambda_{\text{len}}\,|m|/L_{\max}$ to discourage verbose or overly specific guidance, where $|m|$ is the number of memory tokens, $L_{\max}$ is the generation budget, and $\lambda_{\text{len}}>0$ controls the penalty.

\subsection{Decision-Content Decoupled Policy Optimization}
\label{sec:decoupled-optimization}

Applying standard GRPO directly to the structured output $y=d\oplus m$ conflates two distinct learning signals: $d$ governs \emph{whether} memory is generated, while $m$ governs \emph{what} guidance is produced. 
This conflation creates two challenges. First, since Stage~2 is initialized from a supervised policy that favors generation, standard \textit{i.i.d.} sampling may yield groups with no abstain rollouts, eliminating any comparison between generation and abstention. 
Second, the length imbalance between $d$ and $m$ causes content-level updates to dominate the flat per-token objective, suppressing the decision-token gradient.
To address both, we propose \emph{decision-content decoupled policy optimization}, which uses structured counterfactual rollouts to decompose the GRPO advantage into decision- and content-level signals and routes each to the corresponding token positions.

\noindent\textbf{Structured counterfactual rollout.}
For each context $x$, we construct a structured rollout group with one abstain branch and $G-1$ generate branches:
\begin{equation}
y^0=\texttt{[ABSTAIN]}\oplus\varnothing,
\qquad
y^j=\texttt{[GENERATE]}\oplus m^j,\quad j=1,\ldots,G-1.
\label{eq:structured-rollout}
\end{equation}
This guarantees that each group contains both decisions, making the relative value of memory generation versus abstention directly observable.
Since abstention has no guidance content to sample, a single abstain rollout suffices, while the remaining rollouts explore diverse generated memories.

\noindent\textbf{Decision-content advantage decomposition.}
Given the structured rollout group, we decompose the learning signal into a cross-branch decision advantage and a within-branch content advantage:
\begin{equation}
V_{\mathrm{abs}} = R(x,y^0),
\qquad
V_{\mathrm{gen}} = \frac{1}{G-1}\sum_{j=1}^{G-1} R(x,y^j),
\qquad
\Delta = V_{\mathrm{abs}} - V_{\mathrm{gen}} .
\label{eq:branch-values}
\end{equation}
Here, $\Delta$ captures the relative benefit of abstaining over generating memory for the current context.
The \emph{decision} advantage $A_d^j$ uses $\Delta$ as a signed cross-branch signal: $A_d^j = +\Delta$ for the abstain rollout ($j=0$) and $A_d^j = -\Delta$ for generate rollouts ($j\geq 1$). Since $\Delta=V_{\mathrm{abs}}-V_{\mathrm{gen}}$, abstention
receives positive advantage when it outperforms generation, and generation is favored otherwise.

The \emph{content} advantage $A_c^j$ ranks generated memories via group normalization within the generate branch: $A_c^j = 0$ for $j=0$, and $A_c^j = \big(R(x,y^j)-V_{\mathrm{gen}}\big) / \big(\operatorname{std}(\mathbf{r}_{\mathrm{gen}})+\epsilon_{\mathrm{std}}\big)$ for $j\geq 1$, where $\mathbf{r}_{\mathrm{gen}}=\{R(x,y^j)\}_{j=1}^{G-1}$ denotes the rewards of generate rollouts.
That is, this term reduces to the standard GRPO formulation within the generate rollouts.

\noindent\textbf{Token-level credit assignment.}
To route the decomposed signals to the appropriate token positions, we construct a per-token advantage $A_t^j$.
Let $T_d$ denote the length of the decision prefix.
We assign
\begin{equation}
A_t^j =
\begin{cases}
A_d^j, &  t \leq T_d \\[2pt]
\mathbbm{1}\!\left[\Delta<0\right] A_c^j, & t > T_d 
\end{cases}
\label{eq:per-token-advantage}
\end{equation}
Decision tokens receive only the decision-level signal $A_d^j$, while content tokens receive the content-level signal $A_c^j$ only when generation improves over abstention, \ie, $\Delta<0$.
This $\Delta$-gating avoids updating generated content in contexts where memory generation is not beneficial, preventing the assignment of content-level gradients to suboptimal generate decisions.
Substituting $A_t^j$ into the GRPO objective yields the Stage~2 adaptation objective:
\begin{equation}
\mathcal{J}_\text{mem}^{(2)}(\theta)
=
\mathbb{E}_{x}
\left[
\frac{1}{G}
\sum_{j=0}^{G-1}
\frac{1}{|y^j|}
\sum_{t=1}^{|y^j|}
\left(
\min\!\Big(
\rho_t^j A_t^j,\;
\operatorname{clip}\!\big(\rho_t^j,1-\epsilon_{\mathrm{clip}},1+\epsilon_{\mathrm{clip}}\big)A_t^j
\Big)
-\beta D_{\mathrm{KL}}^{(t)}
\right)
\right]
\label{eq:stage2-objective}
\end{equation}
Compared with standard GRPO (Eq.~\ref{eq:grpo-objective}), the only objective-level change is replacing the scalar group-relative advantage $\hat{A}^j$ with the per-token advantage $A_t^j$.
This preserves the GRPO framework while separating the two learning problems: decision tokens learn \emph{when} to generate through cross-branch comparison, and content tokens learn \emph{what} to generate through within-branch ranking.

\section{Experiments}
\label{sec:experiments}

\noindent\textbf{Benchmarks.}
We evaluate on four agentic benchmarks.
\textbf{\textsc{WebArena}}~\citep{zhou2023webarena} contains 812 multi-step browser tasks over five domains (Shopping, CMS, GitLab, Reddit, Maps). Following WebAgent-R1~\citep{wei-etal-2025-webagent} and WebRL~\citep{qi2024webrl}, we use a 647/165 train/test split.
\textbf{\textsc{WorkArena}}~\citep{drouin2024workarena} is an enterprise software web-navigation benchmark on the ServiceNow platform~\citep{servicenow2023vancouver}, covering 33 task templates across four categories (Menu, Form, List, Knowledge). We use 20 seeds per template for training and 10 disjoint seeds for evaluation.
\textbf{\textsc{LifelongAgentBench} (\textsc{LAB})}~\citep{zheng2025lifelongagentbench} tests experience reuse in terminal environments. Following MemRL~\citep{zhang2026memrl}, we use the Database (DB, 22 SQL skills) and Operating System (OS, 29 Bash skills) subsets, each with 500 tasks and a 7:3 split.
\textbf{\textsc{ALFWorld}}~\citep{shridhar2020alfworld} consists of text-based embodied household tasks across six manipulation types. We follow the official split with 3{,}553 train and 134 unseen test tasks.
We use task success rate (SR) as the reward signal across all benchmarks.
We construct the offline experience bank using JEF-Hinter~\citep{nekoei2025just}, which distills raw interaction traces into compact, reusable hints by identifying decisive steps in long trajectories. We emphasize that our \ourmethod{} framework is source-agnostic. Any retrieval-based memory bank, including human demonstrations, agent traces, or curated documentation, can serve as supervision for \mempolicy{}, effectively converting retrieval-based memory into a generative one.

\noindent\textbf{Baselines.}
Beyond the base agents (with no memory), we compare against two memory paradigms.
\emph{(i)~Workflow-based memory}: 
\textbf{RAG}~\citep{lewis2020retrieval} retrieves the top-$k$ experiences from the JEF-Hinter~\citep{nekoei2025just} memory bank via BM25~\citep{robertson2009probabilistic}, effectively matching the approach used in JEF-Hinter. 
\textbf{Mem0}~\citep{chhikara2025mem0} combines RAG with rule-based management. In both settings, we fix $k = 1$.
\emph{(ii) Learning-based memory}: \textbf{Memory-R1}~\citep{yan2025memory} trains a memory manager with outcome-driven RL for structured memory operations. \textbf{MemRL}~\citep{zhang2026memrl} learns Q-values over episodic memory for utility-aware retrieval.

\noindent\textbf{Agent and memory configuration.}
The memory model \ourmethod{} and the downstream agent are two separate models with independent parameters, even when they share the same backbone architecture.
For a fair comparison with Memory-R1~\citep{yan2025memory}, we adopt the same backbone, \texttt{Qwen-2.5-7B-Instruct}~\citep{yang2024qwen2}, for the memory model  \mempolicy{}, on which we apply \ourmethod{}'s two-stage distillation.
Training-based methods use only training-split tasks and their corresponding JEF-Hinter hints. The same split isolation is applied to the RAG and Mem0 banks. All results are evaluated on held-out test tasks.
To assess cross-agent generalization, we evaluate two downstream agents: (i) a \texttt{Qwen-2.5-7B-Instruct} agent fine-tuned under the same setting as WebAgent-R1~\citep{wei-etal-2025-webagent}, also used during stage-2 adaptation distillation, and (ii) the proprietary \texttt{gpt-5.4-mini}.
Section~\ref{sec:experiments} reports text-only results with \texttt{gpt-5.4-mini} as the base agent. Section~\ref{sec:in-depth-analysis} further reports cross-agent evaluation and visual-input ablations on \textsc{WebArena}. In the visual setting, the memory model receives the initial screenshot and visual grounding extracted by \texttt{gemini-2.5-flash}, using \texttt{Qwen-2.5-VL-7B-Instruct} as the visual backbone.
\emph{Implementation details are in Appendix~\ref{app:experimental-details}.}

\subsection{Main Results}
\label{sec:main-results}

\begin{table*}[t]
\centering
\small
\setlength{\tabcolsep}{3.2pt}
\renewcommand{\arraystretch}{1.12}
    \caption{
        Task success rate (SR \%) across four agent benchmarks with \texttt{gpt-5.4-mini} as the base agent.
        \textbf{Bold}: best per column. \underline{Underline}: second best.
    }
    \label{tab:main-results}
    \resizebox{\textwidth}{!}{%
    \begin{tabular}{@{}l rrrrrr rrrrr rrrrrrr rrr r@{}}
    \toprule
    \multirow{2}{*}{\textbf{Method}}
      & \multicolumn{6}{c}{\textbf{\textsc{WebArena}}}
      & \multicolumn{5}{c}{\textbf{\textsc{WorkArena}}}
      & \multicolumn{7}{c}{\textbf{\textsc{ALFWorld}}}
      & \multicolumn{3}{c}{\textbf{\textsc{LAB}}}
      & \multirow{2}{*}{\textbf{Avg.}} \\
    \cmidrule(lr){2-7} \cmidrule(lr){8-12} \cmidrule(lr){13-19} \cmidrule(lr){20-22}
    & \textsc{Shop} & \textsc{CMS} & \textsc{GL} & \textsc{Red} & \textsc{Map} & \textit{Avg}
      & \textsc{M\&D} & \textsc{Form} & \textsc{Filt} & \textsc{SK} & \textit{Avg}
      & \textsc{P\&P} & \textsc{Exam} & \textsc{Cln} & \textsc{Heat} & \textsc{Cool} & \textsc{P2P} & \textit{Avg}
      & \textsc{DB} & \textsc{OS} & \textit{Avg} & \\
    \midrule
    Base Agent
      & 28.4 & 14.6 & 31.2 & 28.8 & 32.4 & 27.1
      & 31.9 & 55.7 & 11.6 & 76.9 & 42.0
      & 88.3 & 82.7 & 86.1 & 85.0 & 85.5 & 78.8 & 85.3
      & 28.5 & 25.0 & 26.8 & 45.3 \\
    \midrule
    RAG
      & 28.6 & 20.2 & 37.4 & 38.2 & 32.4 & 31.4
      & 33.6 & 60.6 & 9.0 & 78.1 & 42.6
      & 89.1 & 83.4 & 86.8 & 85.5 & 85.8 & 79.6 & 87.1
      & 29.9 & 27.2 & 28.5 & 47.4 \\
    Mem0
      & 29.8 & 22.4 & 38.6 & 36.4 & 32.2 & 31.9
      & 33.9 & 61.8 & 9.5 & 79.1 & 44.1
      & 89.7 & 84.1 & 87.3 & 85.8 & 87.2 & 80.5 & 87.5
      & 31.9 & 28.1 & 30.0 & 48.4 \\
    \midrule
    Memory-R1
      & 30.6 & 24.8 & 40.2 & 38.8 & 31.6 & 33.2
      & 35.4 & 62.3 & 10.5 & 80.5 & 44.3
      & 89.9 & 85.0 & 87.7 & 86.3 & 87.7 & 81.1 & 87.9
      & 32.6 & 29.8 & 31.2 & 49.2 \\
    MemRL
      & 31.2 & 26.4 & 41.8 & 40.0 & 30.8 & 34.0
      & 36.5 & 63.6 & 10.6 & 80.5 & 46.1
      & 90.8 & 85.5 & 88.7 & 86.5 & 87.1 & 81.2 & 88.0
      & 33.0 & 30.7 & 31.9 & 50.0 \\
    \midrule
    \ourmethod{}~(Stage~1)
      & \underline{30.6} & \underline{17.4} & \underline{46.8} & \underline{47.8} & \underline{32.4} & \underline{35.0}
      & \underline{41.6} & \underline{65.8} & \underline{9.0} & \underline{81.5} & \underline{46.6}
      & \underline{92.1} & \underline{88.0} & \underline{91.2} & \underline{89.0} & \underline{90.0} & \underline{84.1} & \underline{90.0}
      & \underline{35.3} & \underline{32.9} & \underline{34.1} & \underline{51.4} \\
    \rowcolor{bestcell}
    \ourmethod{}
      & \textbf{34.6} & \textbf{42.8} & \textbf{50.2} & \textbf{52.6} & \textbf{35.4} & \textbf{43.1}
      & \textbf{45.1} & \textbf{70.6} & \textbf{13.6} & \textbf{85.3} & \textbf{50.3}
      & \textbf{94.2} & \textbf{90.2} & \textbf{92.3} & \textbf{91.5} & \textbf{91.6} & \textbf{86.7} & \textbf{91.6}
      & \textbf{38.4} & \textbf{35.0} & \textbf{36.7} & \textbf{55.4} \\
    \bottomrule
    \end{tabular}
    }
    \vspace{-5mm}
\end{table*}

\textbf{\ourmethod{} achieves state-of-the-art performance across all benchmarks and sub-domains.}
As summarized in Table~\ref{tab:main-results},
\ourmethod{} leads every \textsc{WebArena} sub-domain, with the largest absolute gains in Reddit ($+$23.8\,pp) and CMS ($+$28.2\,pp), where structured navigation patterns benefit most from memorized experience.
On \textsc{WorkArena}, \ourmethod{} improves the base agent from 42.0\% to 50.3\% on average, with strong gains on Form ($+$14.9\,pp).
On \textsc{ALFWorld}, \ourmethod{} achieves 91.6\%, a $+$6.3\,pp improvement over the already-strong \texttt{GPT-5.4-mini} baseline.

\textbf{Experience distillation alone already matches or surpasses RL-based baselines.}
\ourmethod{}~(Stage~1) achieves 35.0\% on \textsc{WebArena}, comparable to Memory-R1 (33.2\%) and MemRL (34.0\%) without any RL training.
This validates offline parametric knowledge as a strong initialization strategy.

\textbf{The RL stage provides significant additional gains on WebArena.}
Moving from Stage~1 to the full model yields $+$8.1\,pp on \textsc{WebArena} overall, with the largest jumps on CMS ($+$25.4\,pp), Reddit ($+$4.8\,pp), and Maps ($+$3.0\,pp).
\textsc{ALFWorld} gains a more modest $+$1.6\,pp, consistent with the frontier agent's high baseline leaving less room for improvement.

\subsection{Ablation Study}
\label{sec:ablation}

\noindent\textbf{RQ1: Are both training stages necessary?}\
We compare \ourmethod{} against two single-stage variants.
\textit{(i) w/o Stage~1 init} skips the experience distillation (SFT phase) and applies online RL directly to \texttt{Qwen2.5-7B-Instruct}.
\textit{(ii) Unified single-stage} collapses both stages into one RL phase that jointly optimizes the downstream task reward $R_{\text{task}}$, the same length regularizer $R_m$ used in \ourmethod{}, and an additional BERTScore-based similarity reward $R_{\text{sim}}$~\citep{zhang2019bertscore} computed between the generated memory and the corresponding reference guidance from the training bank, so that the single RL stage has both an imitation signal toward reference hints and a downstream task signal.

\begin{wrapfigure}[28]{r}{0.45\linewidth}
\vspace{-9pt}
\centering

\small
\setlength{\tabcolsep}{2.5pt}
\renewcommand{\arraystretch}{1.1}
\captionof{table}{Ablation results (SR \%) on \textsc{WebArena} and \textsc{ALFWorld}. Subscripts show drop from full model.}
\label{tab:ablation-results}
\begin{tabular}{@{}l cc@{}}
\toprule
\textbf{Variant} & \textbf{\textsc{WA}} & \textbf{\textsc{ALF}} \\
\midrule
\rowcolor{bestcell}
\ourmethod{} (Full) & \textbf{43.1} & \textbf{91.6} \\
\midrule
\multicolumn{3}{@{}l}{\textit{Training Stages}} \\
\textit{w/o} Stage~1 init
    & $37.9_{\scriptscriptstyle-5.2}$
    & $86.9_{\scriptscriptstyle-4.7}$ \\
Unified single-stage
    & $36.3_{\scriptscriptstyle-6.8}$
    & $85.7_{\scriptscriptstyle-5.9}$ \\
\midrule
\multicolumn{3}{@{}l}{\textit{RL Objective}} \\
\textit{w/o} structured rollout
    & $38.3_{\scriptscriptstyle-4.8}$
    & $87.1_{\scriptscriptstyle-4.5}$ \\
\textit{w/o} $\Delta$-gating
    & $41.3_{\scriptscriptstyle-1.8}$
    & $89.6_{\scriptscriptstyle-2.0}$ \\
\textit{w/o} $R_m$
    & $41.9_{\scriptscriptstyle-1.2}$
    & $90.5_{\scriptscriptstyle-1.1}$ \\
\bottomrule
\end{tabular}

\vspace{8pt}
%
\includegraphics[width=\linewidth]{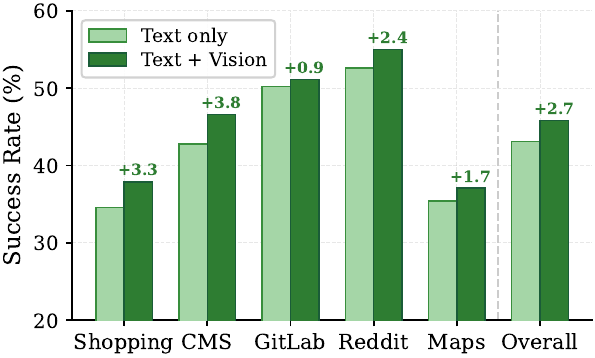}
\captionof{figure}{Performance of \ourmethod{} with and without visual observations on \textsc{WebArena}.}
\label{fig:multimodal}

\vspace{-2pt}
\end{wrapfigure}

Results in Table~\ref{tab:ablation-results} show that \noindent\textbf{both training stages are essential, with unified training suffering the largest drop.}
Removing Stage~1 initialization degrades \textsc{WebArena} by 5.2\,pp, suggesting that without a well-initialized memory distribution, online RL struggles to converge.
Unified single-stage training incurs a larger drop ($-6.8$\,pp on \textsc{WebArena}), indicating that jointly optimizing the imitation reward $R_{\text{sim}}$ and $R_{\text{task}}$ cannot match \ourmethod{}'s staged optimization.
We attribute this to a mismatch between the two rewards: $R_{\text{sim}}$ encourages imitation of reference memories, whereas $R_{\text{task}}$ rewards memories that improve task success. Since useful memories for new tasks may differ from the references, optimizing both rewards in a single stage can produce conflicting gradients.

\noindent\textbf{RQ2: Does Stage-2 decision--content policy optimization help?}\
We design three variants targeting its individual components.
\textit{(i) w/o structured rollout} reverts to vanilla GRPO without paired abstain--generate branches;
\textit{(ii) w/o $\Delta$-gating} replaces gated fusion in Eq.~\ref{eq:per-token-advantage} with a naive sum of decision- and content-level advantages;
\textit{(iii) w/o $R_m$} drops the length-aware memory-quality reward.
Variant \textit{(i)} shows that  \noindent\textbf{structured counterfactual sampling is the most critical Stage-2 component.}
Removing structured rollout costs 4.8\,pp on \textsc{WebArena} and 4.5\,pp on \textsc{ALFWorld}, the largest drops among RL-objective ablations.
Without paired generate-abstain comparisons, most sampling groups lack a routing signal, limiting the policy’s ability to learn \emph{when} to generate memory.
Variants \textit{(ii)} and \textit{(iii)} show that \noindent\textbf{$\Delta$-gating and $R_m$ contribute complementary signals.}
Removing $\Delta$-gating causes consistent drops on both \textsc{WebArena} ($-1.8$\,pp) and \textsc{ALFWorld} ($-2.0$\,pp), suggesting that naive fusion dilutes the optimization signal by allowing content-level updates even when abstention is preferable. Removing $R_m$ leads to mild degradation, potentially indicating that the length regularizer encourages concise memory and helps reduce noise in generated guidance.

\noindent\textbf{RQ3: Do visual observations improve memory generation?}\
In addition to textual states, the web nativation tasks provide multimodal observations such as screenshots.
We therefore build a vision-language variant of \ourmethod{} using \texttt{Qwen2.5-VL-7B-Instruct} as the memory-policy backbone, and compare it with a text-only variant, \ie, \texttt{Qwen2.5-7B-Instruct}, on \textsc{WebArena} sub-domains.

Our experiments show that \noindent\textbf{visual observations provide consistent gains, with larger benefits in visually grounded domains.}
Figure~\ref{fig:multimodal} shows that the multimodal variant outperforms its text-only counterpart across all \textsc{WebArena} sub-domains, improving overall SR by 2.7\,pp.
The gain is largest in CMS ($+3.8$\,pp) and Shopping ($+3.3$\,pp), where page layout and product images provide grounding signals difficult to capture from text alone.
In contrast, GitLab shows the smallest gain ($+0.9$\,pp), consistent with its code-centric content where visual rendering provides limited additional signal.

\subsection{In-Depth Analysis}
\label{sec:in-depth-analysis}

\noindent\textbf{RQ4: How does adaptive abstention improve performance?}\
We further analyze how adaptive abstention contributes to performance gains by examining the relationship between the model's abstention behavior and task difficulty on the \textsc{WebArena} benchmark.
Tasks are grouped into five equal-width bins according to the base agent's success rate, where a lower success rate indicates higher task difficulty.
For each bin, we report \ourmethod{}'s average abstention rate and the corresponding success-rate improvement over the base agent.
\begin{wrapfigure}[42]{r}{0.45\linewidth}
\vspace{-6pt}
\centering

\includegraphics[width=\linewidth]{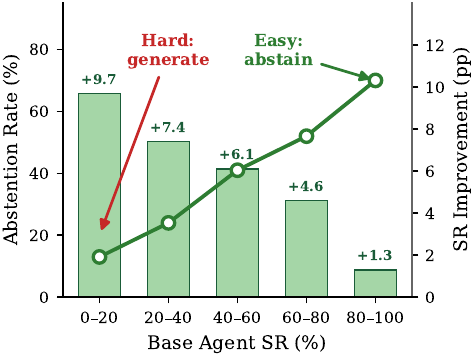}
\captionof{figure}{Abstention rate (line, left axis) and SR improvement over the base agent (bars, right axis) across task-difficulty bins on \texttt{WebArena}.}
\label{fig:adaptivity}

\vspace{6pt}

\small
\setlength{\tabcolsep}{2.5pt}
\renewcommand{\arraystretch}{1.1}
\captionof{table}{Cross-agent transfer (SR \%). Subscripts show gain over base agent.}
\label{tab:cross-agent}
\resizebox{\linewidth}{!}{%
\begin{tabular}{@{}l cc cc@{}}
\toprule
\multirow{2}{*}{\textbf{Method}}
  & \multicolumn{2}{c}{\texttt{Qwen2.5-7B}}
  & \multicolumn{2}{c}{\texttt{GPT-5.4-mini}} \\
\cmidrule(lr){2-3} \cmidrule(lr){4-5}
  & \textbf{\textsc{WA}} & \textbf{\textsc{ALF}}
  & \textbf{\textsc{WA}} & \textbf{\textsc{ALF}} \\
\midrule
Base Agent
    & $27.9\vphantom{0_{\scriptscriptstyle+00.0}}$
    & $72.9\vphantom{0_{\scriptscriptstyle+00.0}}$
    & $27.1\vphantom{0_{\scriptscriptstyle+00.0}}$
    & $85.3\vphantom{0_{\scriptscriptstyle+00.0}}$ \\
+ RAG
    & $32.1_{\scriptscriptstyle+\phantom{0}4.2}$
    & $75.0_{\scriptscriptstyle+\phantom{0}2.1}$
    & $31.4_{\scriptscriptstyle+\phantom{0}4.3}$
    & $87.1_{\scriptscriptstyle+\phantom{0}1.8}$ \\
\rowcolor{bestcell}
+ \ourmethod{}
    & $\mathbf{46.1}_{\scriptscriptstyle+18.2}$
    & $\mathbf{84.7}_{\scriptscriptstyle+11.8}$
    & $\mathbf{43.1}_{\scriptscriptstyle+16.0}$
    & $\mathbf{91.6}_{\scriptscriptstyle+\phantom{0}6.3}$ \\
\bottomrule
\end{tabular}%
}

\vspace{6pt}

\includegraphics[width=\linewidth]{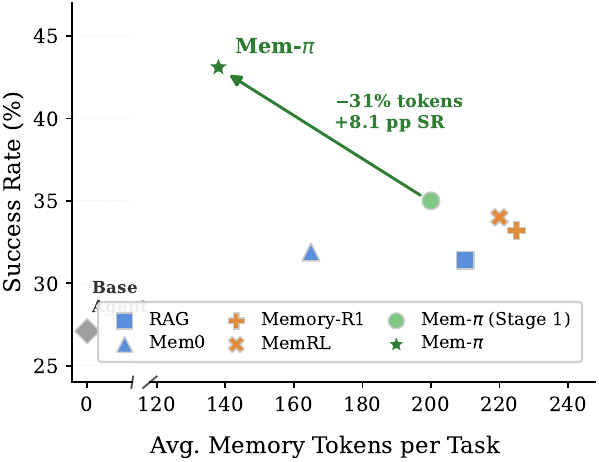}
\captionof{figure}{Performance vs.\ memory-token usage on \textsc{WebArena} across different methods.}
\label{fig:efficiency}
\vspace{-6pt}
\end{wrapfigure}
\noindent\textbf{\ourmethod{} abstains on easy tasks, generates on hard tasks, and improves performance most where memory is needed.}
As shown in Figure~\ref{fig:adaptivity}, abstention is strongly correlated with task difficulty.
On the easiest tasks (base-agent SR 80--100\%), \ourmethod{} abstains in approximately 71\% of cases, while on the hardest tasks, the abstention rate drops to around 13\%.

The SR-improvement bars show the complementary trend: improvement peaks at $+9.7$\,pp on the hardest bin and drops to $+1.3$\,pp on the easiest bin.
This suggests calibrated rather than conservative abstention: \ourmethod{} avoids unnecessary generation when the base agent is already likely to succeed, while generating memory when it provides meaningful benefit.

\noindent\textbf{RQ5: Does adaptive memory generalize across agents?}\
We compare RAG and \ourmethod{} using the training-time agent (\texttt{Qwen2.5-7B-Instruct}) and an unseen frontier agent (\texttt{GPT-5.4-mini}), without retraining the memory model.

\noindent\textbf{\ourmethod{} generalizes to the unseen \texttt{GPT-5.4-mini}, though gains shrink as the base agent becomes stronger.}
As shown in Table~\ref{tab:cross-agent}, \ourmethod{} achieves the strongest transfer on the training-time agent (\texttt{Qwen2.5-7B-Instruct} on \textsc{WebArena}: $+18.2$ vs.\ $+4.2$\,pp over RAG), while retaining a clear advantage with \texttt{GPT-5.4-mini} near the capability ceiling (\textsc{ALFWorld}: $+6.3$ vs.\ $+1.8$\,pp).
Overall, \ourmethod{} yields $3$--$5\times$ larger gains than RAG.
This suggests that memory learned from a weaker agent can encode task guidance at a sufficiently explicit level to remain useful for stronger unseen agents. Training with even weaker agents may further improve interpretability, but could also make Stage~2 RL rewards sparser, introducing a trade-off for future exploration.

\noindent\textbf{RQ6: Is adaptive generative memory token-efficient?}\
We compare task success rate against the average number of memory tokens prepended to the agent input.
For external-memory baselines, this is the length of the retrieved memory content; for generative-memory methods, it is the length of the generated memory. For \ourmethod{}, abstention contributes zero tokens.

\noindent\textbf{\ourmethod{} achieves the best performance-efficiency tradeoff.}
Figure~\ref{fig:efficiency} shows that \ourmethod{} uses 138 memory tokens per task on average, 31\% fewer than Stage~1 (200 tokens) and 38\% fewer than Memory-R1 (225 tokens), while also attaining the highest success rate, improving Stage~1 from 35.0\% to 43.1\%.
This shows that always generating memory is not only inefficient but can be counterproductive: unnecessary memory adds noise to already solvable tasks.
By learning when to abstain, \ourmethod{} reduces memory-token overhead and improves task success, yielding a better performance-efficiency tradeoff.

\begin{figure}[t]
\centering
\setlength{\belowcaptionskip}{-2pt}
\begin{minipage}[c]{0.30\linewidth}
    \centering
    \includegraphics[width=\linewidth]{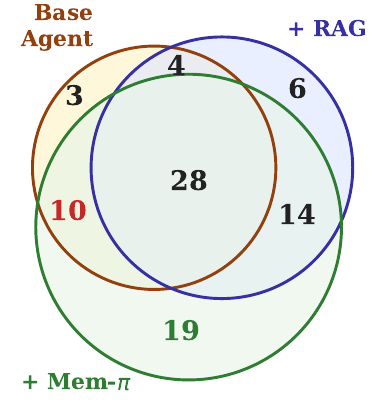}
\end{minipage}%
\hfill
\begin{minipage}[c]{0.68\linewidth}
\begin{tcolorbox}[
    enhanced, skin=enhanced,
    colback=white, colframe=csframe,
    boxrule=0.5pt, arc=2pt,
    left=1pt, right=1pt, top=1pt, bottom=1pt,
]
\fontsize{6.5}{8}\selectfont
\setlength{\parskip}{0pt plus 0.2pt}
\linespread{0.88}\selectfont

\textcolor{darkgreen}{\textbf{Case 1.}}\
\textit{Top search terms (CMS)}.\\
\textbf{Task:} ``List the top 3 search terms in my store.''\\
\textbf{RAG:}\ \textcolor{red!70!black}{\ding{55}}\
``\ldots locate the `Top Search Terms' table\ldots read the \emph{first two rows} and report each `Search Term' with its `Uses' count.''\\
\textbf{\ourmethod{}:}\ \textcolor{darkgreen}{\ding{51}}\
``\ldots take the \emph{first three rows} from the `Search Term' column and output just those term names\ldots''\\
\textbf{Why:} The retrieved hint comes from a top-$2$ source task and transports its ``$2$'' verbatim. Generation conditions on the explicit ``$3$'' in the query and rewrites the count.

\vspace{1pt}\hrule\vspace{1pt}

\textcolor{red!70!black}{\textbf{Case 2.}}\
\textit{Storage option search (Shopping)}.\\
\textbf{Task:} ``\ldots find the best storage option to fit all 40 [Switch game] cards.''\\
\textbf{RAG:}\ \textcolor{red!70!black}{\ding{55}}\
``\ldots search\ldots for `game card case' or `cartridge holder', open listings that show 40+ slots\ldots''\\
\textbf{\ourmethod{}:}\ \textcolor{darkgreen}{\ding{51}}\
$[\texttt{ABSTAIN}]$\\
\textbf{Why:} The query asks for the \emph{best} storage option. The retrieved hint locks search to a narrow product family that excludes broader organizers. Abstention restores the base agent's open search.

\end{tcolorbox}
\end{minipage}
\caption{
    Case analysis of \ourmethod{}, RAG, and the base agent. 
    Left: success-set Venn diagram across the three methods, where the colored numbers mark two \ourmethod{} win patterns over RAG: 19 tasks solved only by \ourmethod{}, and 10 solved by both the base agent and \ourmethod{} but not RAG.
    Right: one sampled case for each pattern, comparing \ourmethod{} and RAG under the same task query.
}
\label{fig:gen-vs-retrieval}
\vspace{-5mm}
\end{figure}

\noindent\textbf{RQ7: Why does adaptive memory outperform retrieval-based memory?}\
We conduct case analysis of the base agent, RAG, and \ourmethod{} on the \textsc{WebArena}. Figure~\ref{fig:gen-vs-retrieval} presents a success-set Venn diagram over these methods and highlights two patterns where \ourmethod{} succeeds and RAG fails.

\noindent\textbf{Generation adapts retrieved guidance to the current query.}
When a query specifies a count, identifier, or format that differs from the most similar memory bank entry, retrieval copies the source verbatim, while generation rewrites it from parametric knowledge to match the current input. Case~1 in Figure~\ref{fig:gen-vs-retrieval} illustrates this: the retrieved source is a top-$2$ task while the query is top-$3$, so RAG's hint reads ``read the first two rows'', whereas \ourmethod{} conditions on the ``$3$'' and produces ``take the first three rows''. Generation resolves the mismatch by rewriting numbers, keys, and formats to fit the query rather than copying from the stored example.

\noindent\textbf{Abstention discards guidance that no longer applies.}
When a retrieved hint encodes assumptions, such as a specific product family or identifier, that do not transfer to the current query, any non-empty hint inherits this bias and misleads the agent. Case~2 in Figure~\ref{fig:gen-vs-retrieval} illustrates this: the retrieved source narrows the search to ``game card case'', but the query asks for the \emph{best} storage option that fits $40$ cards. \ourmethod{} emits \texttt{[ABSTAIN]}, letting the base agent search broadly, while injecting the narrowed hint would have caused failure.
Additional cases are provided in Appendix~\ref{app:case-more}.

\section{Related Work}
\label{sec:related-work}

\noindent \textbf{Learning-based agent memory.}\
Agent memory systems have evolved from static pipelines~\citep{packer2023memgpt,shinn2023reflexion,wang2023voyager} toward learned memory operations jointly optimized with downstream task outcomes~\citep{zhang2024survey,hu2025memory,huang2026rethinking}.
One line of work distills raw interaction trajectories into structured knowledge including rules, guidelines, or strategies that are retrieved at inference time~\citep{zhao2024expel,wang2025agent,wu2025evolver,yang2026autoskill,yang2026plugmem}.
For example, AutoGuide~\citep{fu2024autoguide} compresses offline interaction logs into concise, context-conditional guidelines for web navigation, while ReasoningBank~\citep{ouyang2025reasoningbank} distills generalizable reasoning strategies from both successes and failures, enabling memory-aware test-time scaling.
Another line of work optimizes memory operations end-to-end through reinforcement learning, training controllers to manage storage, retrieval, update, and deletion~\citep{yu2026agentic,yu2025memagent,zhang2026learning,zhang2026memrl,zhan2025exgrpo}, or co-evolving memory architectures with the agent's policy~\citep{xu2025mem,zhou2025memento,xiao2026ui,wang2026memex}.
Memory-R1~\citep{yan2025memory} and Mem-$\alpha$~\citep{wang2025mem} equips an LLM with a dedicated manager that learns structured memory operations through outcome-driven RL.
More recently, SkillRL~\citep{xia2026skillrl} builds a hierarchical skill bank that co-evolves through recursive evolution, and MemEvolve~\citep{zhang2025memevolve} goes further by meta-evolving the memory \emph{architecture} itself across task distributions.
Despite these advances, all remain retrieval-centric: they improve \emph{when} and \emph{how} to access stored entries, but the memory content itself is fixed at write time.
\ourmethod{} departs from this paradigm by modeling memory as a generative policy that dynamically constructs task-adaptive guidance from parametric knowledge.

\noindent \textbf{Generative memory.}\
Compared with retrieval-based memory that stores and retrieves external entries, generative memory encodes experience into model parameters and produces useful information on demand.
One line of work develops parametric memory modules that internalize retrieval behavior into learnable parameters~\citep{qian2024memorag,liu2025memverse,cheng2026conditional,ding2026meki,jaiswal2026memoryllm}.
Early studies compress long contexts~\citep{choi2022prompt,chevalier-etal-2023-adapting,li2024prompt} or external knowledge~\citep{padmanabhan2024propagating,wang2024self} into model parameters, while more recent methods such as MemoryDecoder~\citep{cao2025memory} and MLP Memory~\citep{wei2025mlp} train lightweight networks to imitate non-parametric retrievers.
CoMEM~\citep{wu2025towards} further extends this direction to multimodal settings, showing that a vision-language model can serve as its own memory encoder.
R$^3$Mem~\citep{wang-etal-2025-r3mem} bridges memory retention and retrieval through reversible context compression, improving the faithfulness and usability of parametric memory.
In agent settings, ParamMem~\citep{yao2026parammem} encodes cross-sample reflection patterns into model parameters, enabling diverse and transferable reflection generation.
MemGen~\citep{zhang2025memgen} constructs latent token sequences as machine-native memory through a learned memory trigger and weaver.
Most closely related, SEAM~\citep{li2026beyond} trains a experience adapter with GRPO to generate utility-optimized experience entries for a frozen executor.
These methods show the promise of parameterized memory generation, but largely treat generation as retrieval imitation or an always-on auxiliary process.
\ourmethod{} extends this direction by modeling memory as an adaptive generative policy for multi-step agent interactions, learning both \emph{when} to generate guidance and \emph{what} guidance to generate from downstream agent outcomes.

\section{Conclusion}
\label{sec:conclusion}

We presented \ourmethod{}, a framework that formulates agent memory as a generative policy $\pi_\text{mem}$ rather than retrieval over explicit memory entries.
Through experience distillation and adaptation distillation, \ourmethod{} internalizes reusable behavioral knowledge into model parameters and further refines it with downstream task rewards.
To jointly optimize \emph{when} to generate memory and \emph{what} guidance to generate, we introduced a decision-content decoupled RL objective that separates routing from content optimization through structured counterfactual advantages and per-token credit assignment.
Experiments across web navigation, terminal tool use, and embodied environments show that adaptive memory generation consistently improves over retrieval-based and prior RL-optimized memory baselines, establishing generative memory as an effective alternative to retrieval-centric memory systems for LLM agents.
Future work can extend \ourmethod{} toward closed-loop memory learning, where agents continuously collect new experiences, update their parametric memory.
Another promising direction is grounded and attributable parametric memory, enabling generated guidance to be traced back to supporting experiences while preserving the flexibility of generative memory.

\bibliographystyle{servicenow}
\bibliography{custom,anthology-1,anthology-2}

\newpage
\appendix

\section{Experimental Details}
\label{app:experimental-details}

\subsection{Benchmark Details}
\label{app:benchmarks}

\noindent\textbf{\textsc{WebArena}.}\
\textsc{WebArena}~\citep{zhou2023webarena} is a realistic web-navigation benchmark consisting of 812 tasks across five fully functional web domains.
\textit{Shopping} (162 tasks) involves e-commerce actions such as product search, filtering by attribute, cart management, and order review.
\textit{CMS} (169 tasks) requires interacting with a content management system to create, edit, and publish posts and pages.
\textit{GitLab} (91 tasks) covers collaborative software development workflows including repository management, issue tracking, merge requests, and code review.
\textit{Reddit} (232 tasks) involves social forum interactions: posting, commenting, voting, searching threads, and moderating content.
\textit{Maps} (158 tasks) requires navigating a map service to search for locations, compute directions, and extract geographic information.
Each task requires an agent to interact with dynamic web interfaces through browser actions such as clicking, typing, searching, navigating across pages, and extracting information.
These tasks are long-horizon and often require maintaining intermediate state, reasoning over web-page observations, and recovering from incorrect actions.
Following WebAgent-R1~\citep{wei-etal-2025-webagent}, we adopt the standard train--test split and evaluate on held-out tasks.

\noindent\textbf{\textsc{WorkArena}.}\
\textsc{WorkArena}~\citep{drouin2024workarena} evaluates agents in enterprise workflow scenarios built on the ServiceNow cloud platform.
The benchmark covers four representative workflow categories:
\textit{Dashboard \& Menu Navigation}---locating information across nested menus and dashboards;
\textit{Enterprise Forms}---filling multi-field structured forms with domain-specific validation;
\textit{List Filter/Sort}---applying complex filters, sorting criteria, and pagination on enterprise data lists;
\textit{Knowledge/Service Base}---querying and navigating knowledge articles and service catalog entries.
\textsc{WorkArena} contains 33 atomic task templates, each instantiatable with different random seeds.
We follow the cross-goal generalization setting of BrowserGym~\citep{chezelles2024browsergym}: 20 seeds per template for training and 10 disjoint seeds for evaluation, testing whether agents can generalize learned interaction patterns to unseen goals within the same workflow family.

\noindent\textbf{\textsc{LifelongAgentBench} (\textsc{LAB}).}\
\textsc{LifelongAgentBench}~\citep{zheng2025lifelongagentbench} is designed to evaluate lifelong learning and experience reuse in interactive terminal-based environments.
It contains 1{,}396 total task instances across three environments: Database (DB), Operating System (OS), and Knowledge Graph (KG).
Following prior work~\citep{zhang2026memrl}, we focus on the DB and OS subsets.

The \textit{DB subset} (500 tasks) evaluates 22 SQL-related skills: basic \texttt{SELECT}, filtering (\texttt{WHERE}), grouping (\texttt{GROUP BY}), sorting (\texttt{ORDER BY}), aggregation (\texttt{COUNT}/\texttt{SUM}/\texttt{AVG}/\texttt{MAX}/\texttt{MIN}), nested subqueries, multi-table \texttt{JOIN}s, set operations (\texttt{UNION}/\texttt{INTERSECT}), and data manipulation (\texttt{INSERT}/\texttt{UPDATE}/\texttt{DELETE}).
Execution results are verified automatically via SQL engine output.

The \textit{OS subset} (500 tasks) evaluates 29 Bash-command skills: file and directory operations (\texttt{ls}, \texttt{cp}, \texttt{mv}, \texttt{find}), permission management (\texttt{chmod}, \texttt{chown}), user and group management (\texttt{useradd}, \texttt{groupmod}), text processing (\texttt{grep}, \texttt{sed}, \texttt{awk}, \texttt{wc}), compression (\texttt{tar}, \texttt{gzip}), process inspection (\texttt{ps}, \texttt{top}, \texttt{kill}), and system monitoring (\texttt{df}, \texttt{du}, \texttt{uptime}).
Correctness is verified by checking final OS state.

For both subsets, we construct a 7:3 train--test split with random seed 42 (350 training / 150 test tasks per subset), evaluating whether \ourmethod{} can distill reusable terminal-interaction experience from training tasks and transfer to unseen tasks within the same tool-use domain.

\noindent\textbf{\textsc{ALFWorld}.}\
\textsc{ALFWorld}~\citep{shridhar2020alfworld} is a text-based embodied household environment aligned with embodied simulator states (ALFRED~\citep{shridhar2020alfred}).
It converts household manipulation tasks into textual observations and actions while preserving long-horizon planning difficulty and partial observability.
The benchmark includes six task types:
\textit{pick-and-place} (find an object and place it at a specified location),
\textit{examine-in-light} (find an object and examine it under a light source),
\textit{clean-and-place} (clean an object in a sink and place it at a target),
\textit{heat-and-place} (heat an object in a microwave and place it at a target),
\textit{cool-and-place} (cool an object in a refrigerator and place it at a target),
and \textit{pick-two-and-place} (find two instances of an object and place them together).
These tasks require agents to locate objects, navigate between receptacles, manipulate object states, and place objects at target locations under partial observability.

The original \textsc{ALFWorld} split contains 3{,}553 training tasks, 140 validation-seen tasks, and 134 validation-unseen tasks.
Following prior work~\citep{zhang2026memrl}, we use the 3{,}553 training tasks as the experience pool for memory distillation and evaluate transfer on the 134 validation-unseen tasks.
This setting tests whether the agent can reuse procedural experience from prior household interactions rather than memorizing specific trajectories.

\subsection{Implementation Details}
\label{app:impl}

\noindent\textbf{Implementation stack.}\
We implement \ourmethod{} on top of PyTorch~\citep{paszke2019pytorch} and the Hugging Face \texttt{transformers} library~\citep{wolf-etal-2020-transformers}. RL training is built on TRL~\citep{vonwerra2020trl}, with rollout generation served by vLLM~\citep{kwon2023efficient}.
\ourmethod{} is initialized from \texttt{Qwen2.5-7B-Instruct}~\citep{yang2024qwen2} for the default text-only setting (used in all main-table results), with the multimodal variant \texttt{Qwen2.5-VL-7B-Instruct}~\citep{yang2024qwen2} reserved for the visual-input ablation in Section~\ref{sec:in-depth-analysis} (RQ3); the base agent is a separately fine-tuned \texttt{Qwen2.5-7B-Instruct} kept frozen throughout memory training. Both models use the standard chat template of their base. \texttt{[ABSTAIN]} and \texttt{[GENERATE]} are added to the vocabulary as special tokens, each initialized with the mean embedding of semantically related existing tokens (\textit{skip}/\textit{bypass} for \texttt{[ABSTAIN]}; \textit{recall}/\textit{retrieve} for \texttt{[GENERATE]}), then averaged to produce a shared initial embedding so the two decision tokens start with near-equal logit values ($\approx 50\%$ abstention probability) at the beginning of Stage~2.

\begin{figure}[!p]
\centering
\setlength{\belowcaptionskip}{-2pt}

\begin{tcolorbox}[
    enhanced,
    colback=white, colframe=csframe,
    boxrule=0.5pt, arc=2pt,
    title=\textbf{\textsc{WebArena} \textnormal{\textperiodcentered\,\textit{shopping\_admin (CMS)}}},
    fonttitle=\small\bfseries,
    coltitle=white, colbacktitle=csframe,
    left=4pt, right=4pt, top=3pt, bottom=3pt,
]
\fontsize{7.2}{8.6}\selectfont
\setlength{\parskip}{1.5pt plus 0.3pt}

\begin{minipage}[t]{0.42\linewidth}
\vspace{0pt}%
\centering
\includegraphics[width=\linewidth]{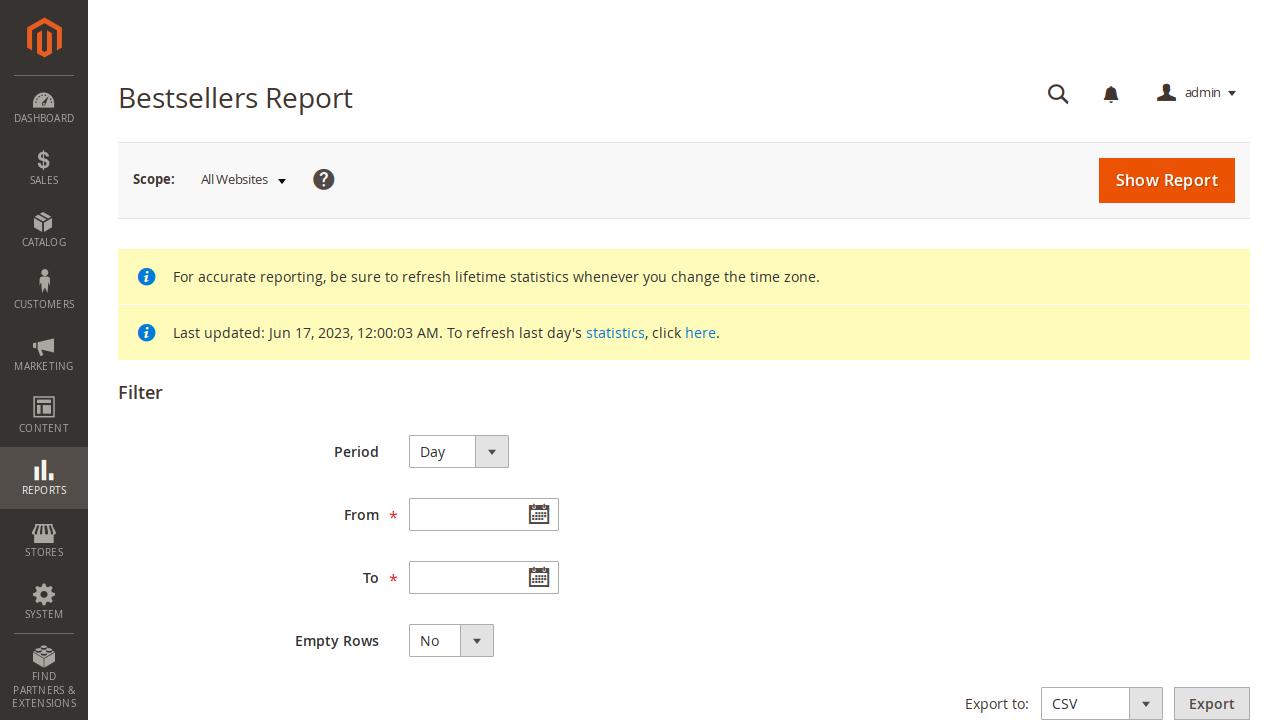}
\end{minipage}\hfill
\begin{minipage}[t]{0.56\linewidth}
\vspace{0pt}%
\textbf{Task query.}\
``What is the top-1 best-selling product in 2022.''

\textbf{Memory (hint).}\
``Use the reports with date filters, not the dashboard widget: go to \textit{Reports}\,$>$\,\textit{Products}\,$>$\,\textit{Bestsellers}, set \textit{Period} to \textit{Year}, enter \textit{From} \texttt{01/01/2022} and \textit{To} \texttt{12/31/2022} (MM/DD/YYYY), confirm scope if needed, click \textit{Show Report}, then sort by \textit{Order Quantity} to get the top-1.''

\textbf{Grounding (excerpt).}\
\textit{Page type}: Magento admin Bestsellers Report. \textit{Key elements}: sidebar with \textit{Reports} selected; main area has a \textit{Period} dropdown (default ``Day''), \textit{From}/\textit{To} date inputs, a \textit{Show Report} button, and informational banners about lifetime-statistics refresh.
\end{minipage}
\end{tcolorbox}

\vspace{1pt}

\begin{tcolorbox}[
    enhanced,
    colback=white, colframe=csframe,
    boxrule=0.5pt, arc=2pt,
    title=\textbf{\textsc{WorkArena} \textnormal{\textperiodcentered\,\textit{filter-incident-list (List)}}},
    fonttitle=\small\bfseries,
    coltitle=white, colbacktitle=csframe,
    left=4pt, right=4pt, top=3pt, bottom=3pt,
]
\fontsize{7.2}{8.6}\selectfont
\setlength{\parskip}{1.5pt plus 0.3pt}

\begin{minipage}[t]{0.42\linewidth}
\vspace{0pt}%
\centering
\includegraphics[width=\linewidth]{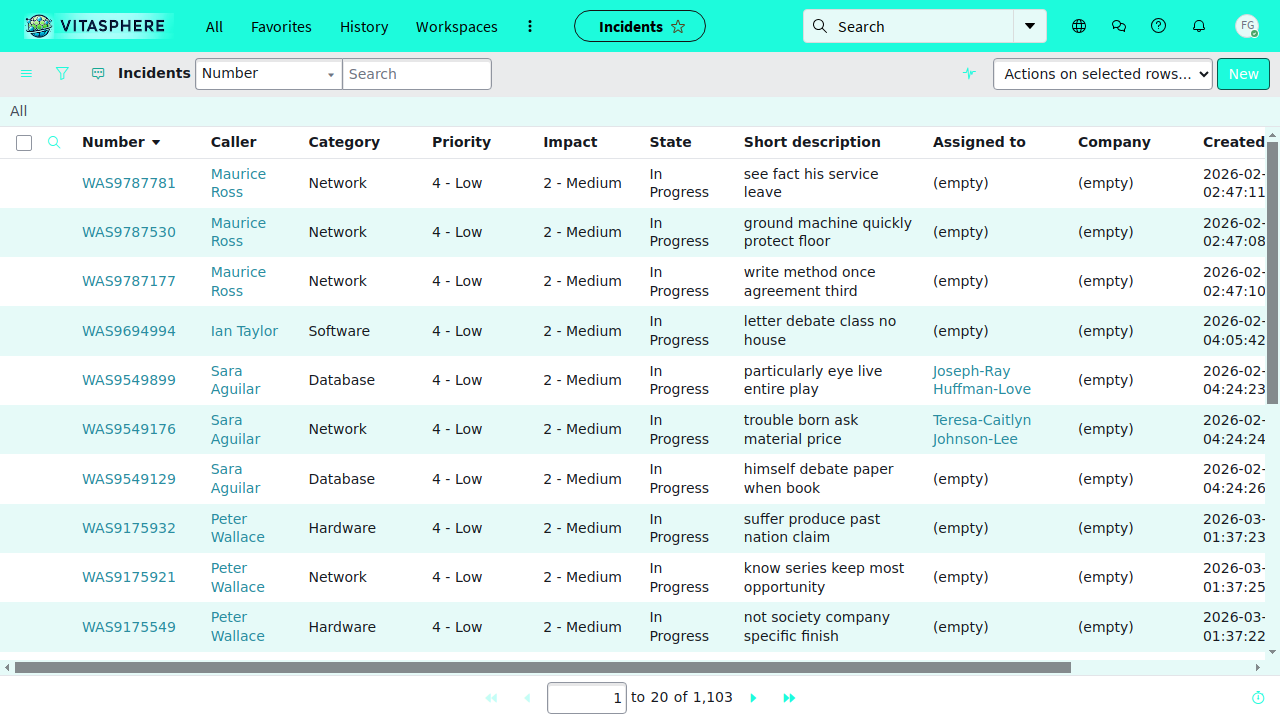}
\end{minipage}\hfill
\begin{minipage}[t]{0.56\linewidth}
\vspace{0pt}%
\textbf{Task query.}\
``Create a filter for the list to extract all entries where \textit{Assigned to} is \textit{ITIL User}, OR \textit{Configuration item} is empty, OR \textit{Short description} is \textit{My desk phone does not work}.''

\textbf{Memory (hint).}\
``Open the filter builder via the \textit{Show\,/\,hide filter} button (not the \textit{All} breadcrumb), set the group combobox to \textit{Any of these conditions must be met} or use \textit{Add OR Condition} on a row, configure each condition (field, operator like \texttt{is}/\texttt{is empty}, value), then click \textit{Run filter} to apply.''

\textbf{Grounding (excerpt).}\
\textit{Page type}: ServiceNow Incident list. \textit{Key elements}: top nav (\textit{All}/\textit{Favorites}/\textit{History}/\textit{Workspaces}/\textit{Incidents}); sortable column headers (\textit{Number}, \textit{Caller}, \textit{Category}, \textit{Priority}, \textit{Impact}, \textit{State}, \textit{Short description}, \textit{Assigned to}, \textit{Company}, \textit{Created}); pagination ``1 to 20 of 4{,}071''; filter / hamburger icons.
\end{minipage}
\end{tcolorbox}

\vspace{1pt}

\begin{tcolorbox}[
    enhanced,
    colback=white, colframe=csframe,
    boxrule=0.5pt, arc=2pt,
    title=\textbf{\textsc{ALFWorld} \textnormal{\textperiodcentered\,\textit{clean-and-place}}},
    fonttitle=\small\bfseries,
    coltitle=white, colbacktitle=csframe,
    left=4pt, right=4pt, top=3pt, bottom=3pt,
]
\fontsize{7.2}{8.6}\selectfont
\setlength{\parskip}{1.5pt plus 0.3pt}

\textbf{Task query.}\
``Place a rinsed plate in the fridge.''

\textbf{Memory (hint).}\
``1.~Look around to find the dirty item you want to clean.\
2.~Go to a suitable cleaning location (sink or basin).\
3.~Clean the item thoroughly using the cleaning location.\
4.~Go to the appropriate storage location (fridge or cabinet).\
5.~Ensure the storage location is open before placing the cleaned item inside.\
\textit{Common pitfall}: forgetting to open the storage location before placing the item leads to failure.''
\end{tcolorbox}

\vspace{1pt}

\begin{tcolorbox}[
    enhanced,
    colback=white, colframe=csframe,
    boxrule=0.5pt, arc=2pt,
    title=\textbf{\textsc{LAB} \textnormal{\textperiodcentered\,\textit{db: column\_alias / os: awk}}},
    fonttitle=\small\bfseries,
    coltitle=white, colbacktitle=csframe,
    left=4pt, right=4pt, top=3pt, bottom=3pt,
]
\fontsize{7.2}{8.6}\selectfont
\setlength{\parskip}{1.5pt plus 0.3pt}

\textbf{Task query (\textsc{LAB-DB}).}\
``Delete patient records where the number of refills left is below the average refills left across all records and the prescription date is before the latest prescription date for the medication \texttt{Aspirin}.''

\textbf{Memory (hint).}\
``1.~Identify the main condition for your \texttt{DELETE} operation and determine whether it requires comparison with aggregate values (\eg, \texttt{AVG} or \texttt{MAX}).\
2.~Use subqueries to compute these aggregates and reference them in the \texttt{WHERE} clause.\
3.~Combine multiple conditions with \texttt{AND}; ensure the subquery on \texttt{MAX(prescription\_date)} is restricted by medication.''

\vspace{1.5pt}\hrule\vspace{1.5pt}

\textbf{Task query (\textsc{LAB-OS}).}\
``Parse \texttt{/var/log/auth.log} to count failed SSH login attempts per user, save the result to \texttt{/report.txt}, and set permissions to read-only for the owner.''

\textbf{Memory (hint).}\
``1.~Verify the input file exists with \texttt{test -f} before piping into \texttt{grep}.\
2.~Use \texttt{grep `Failed password'} to extract failure lines, then \texttt{awk \{print \$NF\}} to take the username column.\
3.~Pipe through \texttt{sort\,|\,uniq -c} to count per user and redirect to \texttt{/report.txt}.\
4.~Apply \texttt{chmod 400 /report.txt} for owner-read-only.''
\end{tcolorbox}

\caption{
    Sample experience entries drawn from the offline bank $\mathcal{E}$ used to train \ourmethod{}, one per benchmark. Each entry contains a task query (\texttt{source\_trace\_goals} in JEF-Hinter~\citep{nekoei2025just}) and the guidance (JEF-Hinter hint) text. For \textsc{WebArena} and \textsc{WorkArena}, the bank additionally stores the initial screenshot and a structured grounding description used by the VL memory variant in Section~\ref{sec:in-depth-analysis}; \textsc{ALFWorld} and \textsc{LAB} are text-only environments and have no visual channel.
}
\label{fig:bank}
\end{figure}

\noindent\textbf{Experience bank.}\
We sample $5$ hints per task with JEF-Hinter~\citep{nekoei2025just} from $\pi_\text{agent}$ trajectories. Each hint is a short procedural summary of the verified flow for one task. To prevent test-set leakage, only hints derived from training tasks (per the train/test splits in Section~\ref{app:benchmarks}) are used as supervised targets in Stage~1 and as references for the BERTScore-based similarity reward $R_{\text{sim}}$ used in the unified-stage ablation variant in Section~\ref{sec:ablation}. Hints derived from test tasks are excluded from training.
Figure~\ref{fig:bank} shows one sample entry per benchmark to illustrate what the bank actually contains: task query, semantic key, hint text, and (for the visual benchmarks) the initial screenshot and grounding text used by the VL variant.

\noindent\textbf{Stage~1: experience distillation.}\
\ourmethod{} is trained to imitate the hints in $\mathcal{E}$ via the autoregressive supervised objective in Eq.~\ref{eq:stage1-objective}.
We use AdamW~\citep{loshchilov2018decoupled} with $\beta_1{=}0.9$, $\beta_2{=}0.999$, learning rate $2{\times}10^{-5}$, cosine schedule with $5\%$ warmup, batch size $32$, weight decay $0.01$, gradient clipping at $1.0$, max sequence length $2{,}048$ tokens, and $3$ epochs.
Stage~1 is parallelized with PyTorch FSDP~\citep{zhao2023pytorch} across $8 \times$ NVIDIA H100-80GB GPUs.

\noindent\textbf{Stage~2: adaptation distillation.}\
Stage~2 fine-tunes \ourmethod{} with the decision-content decoupled GRPO objective in Eq.~\ref{eq:stage2-objective}, implemented as a custom \texttt{GRPOTrainer} subclass on TRL.
For each task we sample a structured rollout (Eq.~\ref{eq:structured-rollout}) of $G{=}4$ branches: one forced \texttt{[ABSTAIN]} (no generation) and three \texttt{[GENERATE]} branches each producing a memory of up to $L_\text{max}{=}256$ tokens at sampling temperature $1.0$ and \texttt{top\_p} $0.95$.
Optimization uses AdamW with learning rate $1{\times}10^{-6}$, $\beta_1{=}0.9$, $\beta_2{=}0.999$, weight decay $0$, batch size $8$ tasks per step, and $200$ optimization steps.
The clip ratio is $\epsilon_\text{clip}{=}0.2$, the KL coefficient is $\beta{=}0.01$, and the advantage normalization constant is $\epsilon_\text{std}{=}10^{-6}$ (Eq.~\ref{eq:grpo-objective}).
The length regularizer $R_m$ uses $\lambda_\text{len}{=}0.1$ with generation budget $L_\text{max}{=}256$ tokens (Eq.~\ref{eq:reward}). The unified-stage ablation variant (Section~\ref{sec:ablation}, RQ1) additionally uses a BERTScore-based similarity reward $R_{\text{sim}}$~\citep{zhang2019bertscore}, computed as the F1 of layer~$12$ \textsc{bert-base-uncased} between the generated memory and the Stage~1 reference hint.
Stage~2 uses DeepSpeed ZeRO-2~\citep{rasley2020deepspeed} with vLLM colocated for inference on the same $8 \times$ H100-80GB GPUs; the policy model and the rollout vLLM instance share GPU memory via vLLM's sleep mode.

\noindent\textbf{Baselines.}\
RAG~\citep{lewis2020retrieval} and Mem0~\citep{chhikara2025mem0} retrieve the top-$k{=}1$ most similar hints from the same $\mathcal{E}$ used by \ourmethod{}; Mem0 additionally applies its rule-based update pipeline. Memory-R1~\citep{yan2025memory} and MemRL~\citep{zhang2026memrl} are run with their official configurations on the same agent and benchmarks.

\noindent\textbf{Evaluation protocol.}\
For \textsc{WebArena}~\citep{zhou2023webarena} and \textsc{WorkArena}~\citep{drouin2024workarena} we use the official benchmark verifiers from BrowserGym; for \textsc{LAB}~\citep{zheng2025lifelongagentbench}, correctness is verified by SQL execution (DB) and OS state checks via the benchmark's built-in verifiers; for \textsc{ALFWorld}~\citep{shridhar2020alfworld}, success is determined by the environment's terminal condition checker. Reported numbers are means over three independent seeds.

\noindent\textbf{Length regularizer scope and saturation.}
The length regularizer $R_m(m){=}{-}\lambda_\text{len}|m|/L_\text{max}$ in Eq.~\ref{eq:reward} is applied only to \texttt{[GENERATE]} branches, where $m\neq\varnothing$. For \texttt{[ABSTAIN]} branches, we set $m{=}\varnothing$ and the reward reduces to $\operatorname{TaskReward}(\agentpolicy{}(\cdot\mid x))$, as in the second case of Eq.~\ref{eq:reward}. Generation is capped by the decoding budget $L_\text{max}{=}256$ via vLLM's \texttt{max\_new\_tokens}, so $|m|\in[1,L_\text{max}]$ for every \texttt{[GENERATE]} branch and $R_m\in[-\lambda_\text{len},0]$. With $\lambda_\text{len}{=}0.1$, the penalty saturates at $-0.1$ and is an order of magnitude smaller than the binary task reward $\operatorname{TaskReward}\in\{0,1\}$. We strip prompt-leakage tokens, including chat-template scaffolding and instruction echoes, before counting $|m|$, so the penalty reflects only substantive memory content. By Eq.~\ref{eq:reward} and Eq.~\ref{eq:structured-rollout}, $R_m$ enters $V_\text{gen}$ and therefore appears in the decision advantage $A_d$ as well as the content advantage $A_m$. The induced systematic bias on $A_d$ equals the average per-task value of $R_m$, bounded above by $\lambda_\text{len}{=}0.1$ and close to $0.05$ at observed memory lengths. This is an order of magnitude below the per-task $\operatorname{TaskReward}\in\{0,1\}$ gap whenever generation flips downstream success, so the routing gradient on $A_d$ remains dominated by task outcome rather than length cost. 


\begin{figure}[!t]
\begin{subfigure}[b]{0.45\linewidth}
    \centering
    \includegraphics[width=\linewidth]{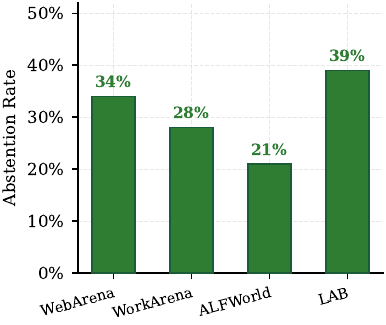}
    \subcaption{Abstention rate per benchmark.}
    \label{fig:abstention-bar}
\end{subfigure}
\hfill
\begin{subfigure}[b]{0.45\linewidth}
    \centering
    \includegraphics[width=\linewidth]{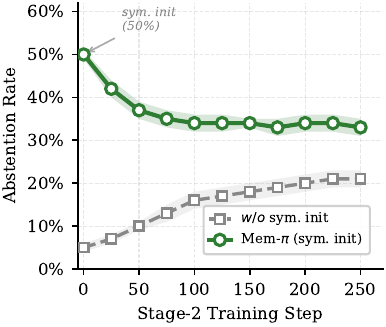}
    \subcaption{Training dynamics on WebArena.}
    \label{fig:abstention-dyn}
\end{subfigure}
\caption{
    \textbf{Abstention statistics.}
    Left: final per-benchmark abstention rates after Stage~2 training.
    Right: abstention rate vs.\ training step, comparing symmetric
    vs.\ standard embedding initialization.
}
\label{fig:abstention-stats}
\vspace{-5mm}
\end{figure}

%

\subsection{Abstention Training Dynamics}
\label{app:abstention-dynamics}

We report per-benchmark final abstention rates and track the evolution of abstention probability during Stage~2 training, starting from the symmetrically initialized Stage~1 checkpoint ($\approx$50\% initial abstention probability).

\noindent\textbf{Abstention rates are benchmark-dependent and converge quickly.}
Final abstention rates vary across benchmarks: 34\% on \textsc{WebArena}, 28\% on \textsc{WorkArena}, 21\% on \textsc{ALFWorld}, and 39\% on \textsc{LAB} (Figure~\ref{fig:abstention-bar}).
The higher rate on \textsc{LAB} reflects a greater proportion of novel configurations where no relevant experience exists; the lower rate on \textsc{ALFWorld} reflects its narrow household-task distribution where memorized patterns apply broadly.

\noindent\textbf{Symmetric initialization is essential for cold-start exploration.}
Figure~\ref{fig:abstention-dyn} shows that with symmetric initialization, the abstention rate converges within approximately 100 training steps to a stable task-appropriate level.
Without symmetric initialization, the \texttt{[ABSTAIN]} token starts with near-zero probability and recovers only slowly to approximately 21\%, never reaching the converged rate of the symmetric variant.
This confirms that balanced initialization of the decision tokens is critical for enabling RL to explore both branches effectively from the start.

\subsection{Additional Case Studies}
\label{app:case-more}


We complement case analysis (Figure~\ref{fig:gen-vs-retrieval}) by examining one representative task per Venn region.
The eight regions partition the test split into qualitatively distinct outcome patterns, summarized below.
\textbf{Region 001} contains \ourmethod{}-only successes, \emph{Pattern 1} of the main text where generation reaches what retrieval cannot.
\textbf{Region 101} contains tasks Base and \ourmethod{} solve but RAG breaks, \emph{Pattern 2} where abstention recovers from RAG noise.
\textbf{Region 011} contains tasks where memory is needed and both RAG and \ourmethod{} succeed.
\textbf{Region 111} contains tasks easy enough for any approach.
\textbf{Region 110} contains \ourmethod{} regressions, where Base and RAG solve but \ourmethod{} fails.
\textbf{Region 010} contains RAG-only successes, where retrieval surfaces a verified procedure that generation re-orders incorrectly.
\textbf{Region 100} contains tasks where any memory hurts and only the unassisted base agent succeeds.
\textbf{Region 000} contains tasks no method solves, typically reflecting environment- or tool-level limitations.
We present a representative case for each region below in the same layout as Figure~\ref{fig:gen-vs-retrieval}.
Long task queries and hints are abridged with ellipses, keeping only the contrastive sub-strings.

\begin{tcolorbox}[
    enhanced, skin=enhanced,
    colback=white, colframe=csframe,
    boxrule=0.5pt, arc=2pt,
    left=4pt, right=4pt, top=2pt, bottom=2pt,
]
\small
\setlength{\parskip}{1pt}

\textbf{Region 001: \ourmethod{} wins Pattern 1 -- Generation reaches what retrieval cannot.} \hfill 15 tasks

\textbf{Case A1 (Task~8).} \textit{Top search terms (Magento admin).}\\
\textbf{Task:} ``List the top 3 search terms in my store.''\\
\textbf{RAG:}\ \textcolor{red!70!black}{\ding{55}}\
``\ldots locate the `Top Search Terms' table\ldots read the \emph{first two rows} and report each `Search Term' with its `Uses' count.''\\
\textbf{\ourmethod{}:}\ \textcolor{darkgreen}{\ding{51}}\
``\ldots take the \emph{first three rows} from the `Search Term' column and output just those term names\ldots''\\
\textbf{Why:} The retrieved hint inherits its source's top-$2$ specification verbatim. Generation conditions on the explicit ``$3$'' in the query and rewrites the count.

\end{tcolorbox}

\begin{tcolorbox}[
    enhanced, skin=enhanced,
    colback=white, colframe=csframe,
    boxrule=0.5pt, arc=2pt,
    left=4pt, right=4pt, top=2pt, bottom=2pt,
]
\small
\setlength{\parskip}{1pt}

\textbf{Region 101: \ourmethod{} wins Pattern 2 -- Abstention recovers from RAG noise.} \hfill 10 tasks

\textbf{Case A2 (Task~41).} \textit{Storage option search (shopping).}\\
\textbf{Task:} ``\ldots find the best storage option to fit all 40 [Switch game] cards.''\\
\textbf{RAG:}\ \textcolor{red!70!black}{\ding{55}}\
``\ldots search\ldots for `game card case' or `cartridge holder', open listings that show 40+ slots\ldots''\\
\textbf{\ourmethod{}:}\ \textcolor{darkgreen}{\ding{51}}\
$[\texttt{ABSTAIN}]$\\
\textbf{Why:} The retrieved hint locks the search to a narrow product family that excludes broader organizers. Abstention restores the base agent's open search, which selects an organizer that better satisfies ``best''.

\end{tcolorbox}

\begin{tcolorbox}[
    enhanced, skin=enhanced,
    colback=white, colframe=csframe,
    boxrule=0.5pt, arc=2pt,
    left=4pt, right=4pt, top=2pt, bottom=2pt,
]
\small
\setlength{\parskip}{1pt}

\textbf{Region 011: Memory is needed, RAG and \ourmethod{} both solve.} \hfill 12 tasks

\textbf{Case A3 (Task~2).} \textit{Review keyword count (Magento admin).}\\
\textbf{Task:} ``Tell me the number of reviews that our store received\ldots that mention term `best'.''\\
\textbf{RAG:}\ \textcolor{darkgreen}{\ding{51}}\
``\ldots `Marketing' > `User Content' > `All Reviews', click `Reset Filter', enter `best' in the `Review' filter\ldots click `Search', then read the `records found' counter\ldots''\\
\textbf{\ourmethod{}:}\ \textcolor{darkgreen}{\ding{51}}\
``\ldots same Magento navigation, enter the keyword in the `Review' filter row\ldots read the `records found' count\ldots''\\
\textbf{Why:} The retrieved entry is exactly aligned with the query, so retrieval and generation converge on the same correct procedure.

\end{tcolorbox}

\begin{tcolorbox}[
    enhanced, skin=enhanced,
    colback=white, colframe=csframe,
    boxrule=0.5pt, arc=2pt,
    left=4pt, right=4pt, top=2pt, bottom=2pt,
]
\small
\setlength{\parskip}{1pt}

\textbf{Region 111: All three solve.} \hfill 28 tasks

\textbf{Case A4 (Task~5).} \textit{Comment-score query (Reddit forum).}\\
\textbf{Task:} ``\ldots count of comments that have received more downvotes than upvotes for the user who made the latest post on the Showerthoughts forum.''\\
\textbf{RAG:}\ \textcolor{darkgreen}{\ding{51}}\
``\ldots set sort to `New', open the latest post, click the author's name, open `Comments', count comments with a leading $-$\ldots''\\
\textbf{\ourmethod{}:}\ \textcolor{darkgreen}{\ding{51}}\
``\ldots open Showerthoughts, sort `New', open top post, click author, `Comments', count comments with negative score\ldots''\\
\textbf{Why:} The agent already follows the standard navigation. Both memory variants reinforce the natural plan rather than introducing new structure.

\end{tcolorbox}

\begin{tcolorbox}[
    enhanced, skin=enhanced,
    colback=white, colframe=csframe,
    boxrule=0.5pt, arc=2pt,
    left=4pt, right=4pt, top=2pt, bottom=2pt,
]
\small
\setlength{\parskip}{1pt}

\textbf{Region 110: \ourmethod{} regresses, Base and RAG solve.} \hfill 4 tasks

\textbf{Case A5 (Task~88).} \textit{Theme preview (Magento admin).}\\
\textbf{Task:} ``Preview the Magento Blank theme for my shop.''\\
\textbf{RAG:}\ \textcolor{darkgreen}{\ding{51}}\
``\ldots find the `Magento Blank' row\ldots click the `View' link \emph{in the Action column}\ldots''\\
\textbf{\ourmethod{}:}\ \textcolor{red!70!black}{\ding{55}}\
``\ldots click the theme name in the grid\ldots open the preview\ldots \emph{avoid clicking the Action column}\ldots''\\
\textbf{Why:} Generation produces a plausible but inverted instruction, telling the agent to avoid the very element that solves the task. Retrieval, drawing on the source task's verified flow, preserves the correct ``Action $\to$ View'' click.

\end{tcolorbox}

\begin{tcolorbox}[
    enhanced, skin=enhanced,
    colback=white, colframe=csframe,
    boxrule=0.5pt, arc=2pt,
    left=4pt, right=4pt, top=2pt, bottom=2pt,
]
\small
\setlength{\parskip}{1pt}

\textbf{Region 010: RAG-only success.} \hfill 6 tasks

\textbf{Case A6 (Task~17).} \textit{Pending review count (Magento admin).}\\
\textbf{Task:} ``What is the total count of Pending reviews amongst all the reviews?''\\
\textbf{RAG:}\ \textcolor{darkgreen}{\ding{51}}\
``\ldots `Marketing' > `User Content' > `Pending Reviews', read the `X records found' label\ldots if unsure about filters, click `Reset Filter'\ldots''\\
\textbf{\ourmethod{}:}\ \textcolor{red!70!black}{\ding{55}}\
``\ldots `Pending Reviews', \emph{click `Reset Filter' to clear any saved filters}, then read the records-found label\ldots''\\
\textbf{Why:} The two hints are nearly identical, but the generated version starts with an unconditional `Reset Filter' click. In our trajectory this clears a server-side default that is needed for the page to populate, while the retrieved hint defers the reset to a later checkpoint. The ordering matters and retrieval preserves the source's correct sequence.

\end{tcolorbox}

\begin{tcolorbox}[
    enhanced, skin=enhanced,
    colback=white, colframe=csframe,
    boxrule=0.5pt, arc=2pt,
    left=4pt, right=4pt, top=2pt, bottom=2pt,
]
\small
\setlength{\parskip}{1pt}

\textbf{Region 100: Both memory methods hurt, only Base solves.} \hfill 3 tasks

\textbf{Case A7 (Task~36).} \textit{Spend aggregation across orders (shopping).}\\
\textbf{Task:} ``How much I spent on food shopping\ldots from mid Jan to the end Jan 2023.''\\
\textbf{RAG:}\ \textcolor{red!70!black}{\ding{55}}\
``\ldots `My Orders'\ldots scroll to the bottom and use the pagination links by their text\ldots open each order from Jan~15--31 via the `View Order' link\ldots sum item subtotals\ldots''\\
\textbf{\ourmethod{}:}\ \textcolor{red!70!black}{\ding{55}}\
``\ldots `My Orders', set `Show per page' to 50\ldots for each order dated 1/15--1/31 click `View Order'\ldots record the `Grand Total' and sum\ldots''\\
\textbf{Why:} Both hints prescribe a multi-page enumeration plus arithmetic that exceeds the agent's effective context window in this trajectory. The unassisted base agent shortcut, summing only the few orders that appear after applying a date filter, happens to be sufficient. Memory hurts here by enforcing a thorough but unsustainable plan.

\end{tcolorbox}

\begin{tcolorbox}[
    enhanced, skin=enhanced,
    colback=white, colframe=csframe,
    boxrule=0.5pt, arc=2pt,
    left=4pt, right=4pt, top=2pt, bottom=2pt,
]
\small
\setlength{\parskip}{1pt}

\textbf{Region 000: No method solves.} \hfill 87 tasks

\textbf{Case A8 (Task~0).} \textit{Bestseller report (Magento admin).}\\
\textbf{Task:} ``What are the top-$3$ best-selling product in Jan 2023.''\\
\textbf{RAG:}\ \textcolor{red!70!black}{\ding{55}}\
``\ldots `Reports > Products > Bestsellers'\ldots `Period' = `Month', `From' 01/01/2023, `To' 01/31/2023\ldots if you see 0 records, refresh stats via the `statistics'/`here' link\ldots''\\
\textbf{\ourmethod{}:}\ \textcolor{red!70!black}{\ding{55}}\
``\ldots same report\ldots if it says `We couldn't find any records', click the `statistics'/`here' link or go to `Reports > Statistics'\ldots''\\
\textbf{Why:} Both hints correctly describe the report path, but the underlying Magento instance does not auto-refresh the bestseller statistics within the agent's step budget. The failure is a tool-level limitation rather than a memory-quality issue. We include this region to show that adaptive memory does not manufacture progress on tasks beyond the agent's tool capabilities.

\end{tcolorbox}

\noindent\textbf{Summary of regions.}
The two highlighted regions tied to \ourmethod{}'s wins over RAG, regions 001 and 101, account for $15 + 10 = 25$ tasks where adaptive generative memory contributes through mechanisms unavailable to retrieval.
The shared-success regions, 011 and 111, contain $12 + 28 = 40$ tasks and confirm that adaptive memory does not displace retrieval when retrieval is well aligned.
The opposing regions, 110, 010, and 100, cost $4 + 6 + 3 = 13$ tasks where \ourmethod{} fails and at least one of the two baselines succeeds. The dominant sub-mode is region 010 where retrieval surfaces a verified procedure that generation re-orders incorrectly.
The all-failure region 000 reflects environment- or tool-level limitations rather than memory quality.


\end{document}